%% file: main.tex
\documentclass{article}
\usepackage{iclr2027_conference,times}

\input{macros}

\title{\penv: Training LLM Agents in Fictional Worlds}

\newcommand{\frontpagethanks}{\thanks{%
Correspondence: \href{mailto:anmol@cs.cornell.edu}{\texttt{anmol@cs.cornell.edu}}. Code: \href{https://github.com/kilian-group/phantom-envs}{\texttt{github.com/kilian-group/phantom-envs}}.
}}

\author{%
Anmol Kabra$^1$\frontpagethanks,
Swathi Saravana Selvam$^1$,
Albert Gong$^1$,
Chao Wan$^1$,\\
\textbf{Christian Belardi$^1$,
Dongyoung Go$^1$,
Katie Z. Luo$^2$,
Kilian Q. Weinberger$^1$}\\
{\normalfont $^1$Cornell University,
$^2$Stanford University}
}

\iclrfinalcopy
\begin{document}

\maketitle
\lhead{Preprint}

\begin{abstract}
Training LLM agents with reinforcement learning (RL) is bottlenecked by environments, which must provide verifiable rewards, support long-horizon interaction, and scale cheaply.
Existing approaches rely on costly human-curated data or on LLM-generated environments that risk hallucinations and benchmark contamination.
We show that LLMs can instead be trained into capable search agents using synthetic environments generated entirely by rules, whose generation requires no LLM and has zero marginal cost.
We build \penv, multi-turn RL environments from fictional worlds, where agents must search a corpus of templated articles to answer multi-hop questions.
Despite sharing no facts with the real world, these strikingly simple environments yield agents that transfer to real-world multi-hop search benchmarks, often outperforming real-world training data on newer benchmarks.
Trained agents generalize to unseen fictional universes, and Qwen models learn to scale their search budget roughly linearly with question difficulty, suggesting emergent search scaling from environment interaction alone.
Ablating environment complexity reveals that hop count drives transfer more than constraints or comparisons: even the simplest rule-generated environments are a surprisingly effective, free resource for training generalizable LLM agents.
\end{abstract}

\input{sections/01_introduction}
\input{sections/04_relatedwork}
\input{sections/02_methods}
\input{sections/03_results}
\input{sections/05_analysis}

\input{sections/05_conclusion}

\subsection*{AI use statement}

In this work, we used generative AI tools for code implementation and experimentation.
We have not used generative AI tools for generating synthetic datasets (the paper is about LLM-free synthetic data), or interpreting results; the rest of the required disclosure tasks are not applicable to this work.
Additionally, we used generative AI tools for brainstorming, editing for readability, and plotting.
We have reviewed all AI-assisted work by verifying code and confirming all literature. We take responsibility for the final content of this work, including text, claims or artifacts produced with the aid of generative AI.

\subsection*{Reproducibility statement}

We use open-source LLMs, training code, and evaluation benchmarks, and our training experiments are across multiple seeds.
All details are in the appendix.
We report standard errors in all tables and plots and note significance.
We have open-sourced our code at \href{https://github.com/kilian-group/phantom-envs}{\texttt{github.com/kilian-group/phantom-envs}}.

\section*{Acknowledgements}

Authors acknowledge help from AI models in various project phases.
DG is supported by Empire AI Postdoctoral Fellowship.
Authors acknowledge compute resources from the National Artificial Intelligence Research Resource (NAIRR) Pilot, Purdue Anvil AI, and NVIDIA’s DGX Station compute platform through their Early Access Program.
This work is supported by the National Science Foundation (NSF) grants RI-2530143, OAC-2118310, IIS-2530143 and through the AI Research Institutes program Award No. DMR-2433348.
This work was partially supported by funding from NewYork-Presbyterian for the NYP-Cornell Cardiovascular AI Collaboration, the National Institute of Food and Agriculture (USDA/NIFA), the Air Force Office of Scientific Research (AFOSR), and a Schmidt AI2050 Senior Fellowship, a Schmidt Sciences program.
We thank anonymous reviewers for their helpful feedback.

\bibliography{refs}
\bibliographystyle{iclr2027_conference}

\appendix
\crefalias{section}{appendix}
\crefalias{subsection}{appendix}
\crefalias{subsubsection}{appendix}
\input{appendix/01_implementation_details.tex}
\input{appendix/02_additional_results.tex}

\end{document}

%% file: macros.tex
\usepackage[utf8]{inputenc} %
\usepackage[T1]{fontenc}    %
\usepackage{hyperref}       %
\usepackage{url}            %
\usepackage{booktabs}       %
\usepackage{amsfonts}       %
\usepackage{nicefrac}       %
\usepackage{microtype}      %
\usepackage[dvipsnames,table]{xcolor}
\usepackage{graphicx}
\usepackage{xspace}
\usepackage{tcolorbox}
\usepackage{listings}
\usepackage{multirow}
\usepackage{subcaption}
\usepackage{enumitem}
\usepackage{mathtools}
\usepackage{wrapfig}

\usepackage[disable]{todonotes}

\usepackage{algorithm}
\usepackage{algpseudocode}
\usepackage{amsmath}
\usepackage{amssymb}
\usepackage{amsthm}

\hypersetup{
	colorlinks,
	linkcolor=MidnightBlue,
	citecolor=MidnightBlue,
	urlcolor=MidnightBlue
}
\usepackage{backref}
\usepackage{cleveref}

\definecolor{myYellow}{HTML}{E8A93C}
\definecolor{myOrange}{HTML}{D96831}
\definecolor{myorange}{HTML}{D96831}
\definecolor{lightorange}{HTML}{FBE5DA}
\definecolor{myGreen}{HTML}{3B9B7B}
\definecolor{myBlue}{HTML}{3B8FBF}
\definecolor{myBrown}{HTML}{8B5A3C}
\definecolor{myDarkGray}{HTML}{A9A9A9}
\definecolor{myPurple}{HTML}{8B5FA8}

\newcommand{\hp}{HotpotQA\xspace}
\newcommand{\twowiki}{2WikiMultihopQA\xspace}
\newcommand{\msq}{MuSiQue\xspace}
\newcommand{\cofca}{CofCA\xspace}
\newcommand{\synthrm}{SynthWorlds-RM\xspace}
\newcommand{\synthsm}{SynthWorlds-SM\xspace}
\newcommand{\frames}{FRAMES\xspace}

\newcommand{\pw}{PhantomWiki\xspace}
\newcommand{\penv}{PhantomEnvironments\xspace}
\newcommand{\penvshort}{PhantomEnvs\xspace}

\newcommand{\hops}{Hops\xspace}
\newcommand{\hopscons}{\hops\unskip+Constraints\xspace}
\newcommand{\hopscomp}{\hops\unskip+Comparisons\xspace}

\newcommand{\gsminf}{GSM-$\infty$\xspace}

\newcommand{\nqhotpot}{NQ+\hp}

\newcommand{\searchrone}{Search-R1\xspace}
\newcommand{\qwenthree}{Qwen2.5-3B-Instruct\xspace}
\newcommand{\qwenseven}{Qwen2.5-7B-Instruct\xspace}
\newcommand{\llamathree}{Llama-3.2-3B-Instruct\xspace}
\newcommand{\phimini}{Phi-4-mini-instruct\xspace}

\newcommand{\answertags}{\texttt{<answer>...</answer>}\xspace}
\newcommand{\searchtags}{\texttt{<search>...</search>}\xspace}
\newcommand{\infotags}{\texttt{<information>...</information>}\xspace}

\newcommand{\inbraces}[1]{\left\{#1\right\}}

\newcommand{\set}[1]{\inbraces{#1}}

\usepackage{listingsutf8} %

\definecolor{yamlkey}{RGB}{0,0,255}      %
\definecolor{yamlvalue}{RGB}{0,128,0}    %
\definecolor{yamlcomment}{RGB}{128,128,128} %
\definecolor{yamlstring}{RGB}{163,21,21}  %

\lstdefinelanguage{yaml}{
  keywords={true,false,null,yes,no},
  keywordstyle=\color{yamlvalue}\bfseries,
  basicstyle=\ttfamily\small,
  sensitive=false,
  comment=[l]{\#},
  commentstyle=\color{yamlcomment}\ttfamily,
  stringstyle=\color{yamlstring}\ttfamily,
}

\lstdefinestyle{yamlstyle}{
  language=yaml,
  backgroundcolor=\color{gray!10},
  frame=single,
  frameround=tttt,
  numbers=left,
  numberstyle=\tiny\color{gray},
  stepnumber=1,
  numbersep=5pt,
  showspaces=false,
  showstringspaces=false,
  showtabs=false,
  tabsize=2,
  captionpos=b,
  breaklines=true,
  breakatwhitespace=false,
  escapeinside={\%*}{*)}
}
\lstdefinestyle{pythonstyle}{
  language=Python,
  basicstyle=\ttfamily\scriptsize,
  keywordstyle=\color{yamlkey}\bfseries,
  commentstyle=\color{yamlcomment}\ttfamily,
  stringstyle=\color{yamlstring}\ttfamily,
  backgroundcolor=\color{gray!10},
  frame=single,
  numbers=left,
  numberstyle=\tiny\color{gray},
  numbersep=5pt,
  showstringspaces=false,
  tabsize=4,
  captionpos=b,
  breaklines=true,
}

%% file: sections/01_introduction.tex
\section{Introduction}
\label{sec:introduction}

Training LLM search agents with reinforcement learning (RL) is bottlenecked by environment curation.
RL needs an environment the agent can interact with: one that yields long multi-turn trajectories, a verifiable reward at the end of each one, and multiple interactions.
Building such environments at scale for RL is expensive.
Wikipedia-grounded environments built from human-curated corpora like NaturalQuestions and \hp require costly annotation and could tie the agent to a fixed Wikipedia snapshot~\citep{kwiatkowski-etal-2019-naturalqa,yang2018hotpotqa,jin2025searchr1}.
LLM-synthesized environments use frontier models to create retrievals, questions, or whole task harnesses~\citep{sun2025zerosearch,gao2025asearcher,wu2026webdancer,goldie2025swirl,databricks2026karl,gandhi2026endless,sullivan2025randomworld}.
They reduce curation effort, but they introduce hallucinated rewards, benchmark contamination risk, API cost, and a capability ceiling set by the generator.
The bottleneck in training LLM agents has shifted from optimization to the question of where verifiable, interactive, long-horizon environments should come from.

We show that such an environment for training LLM agents can be designed \emph{by codified rules alone}, with no humans and no LLMs in the pipeline.
We turn fictional worlds and template-generated multi-hop questions of \pw~\citep{gong2025phantomwiki} into interactive multi-turn environments for RL fine-tuning LLMs, which we call \penv (\penvshort for short).
Such ``rule-generated synthetic environments'' are verifiable by construction and incur zero generation cost.
Universe size, question difficulty, and question budget are knobs the experimenter sets directly, so such environments can scale to arbitrary sizes and complexity.

A priori, there is little reason to expect simple synthetic environments to work.
Fictional universes in these rule-generated synthetic environments share no entities, no facts, and no document distribution with any real-world benchmark.
Articles are template-generated rather than human-written and the relational structure is a randomly sampled social graph---neither the linguistic richness nor the long-tailed topical structure of real-world corpora like Wikipedia.
An agent trained here has no facts to memorize and no exposure to the linguistic variation it will face at evaluation, so the default expectation is that whatever it learns stays coupled to the templates and fails to transfer.
Any transfer that \textit{does} occur due to \penv must come from learning \textit{generalizable agentic search skills}---decomposing questions, retrieving documents correctly, composing knowledge across turns.

\begin{figure}[t]
    \centering
    \includegraphics[width=\linewidth]{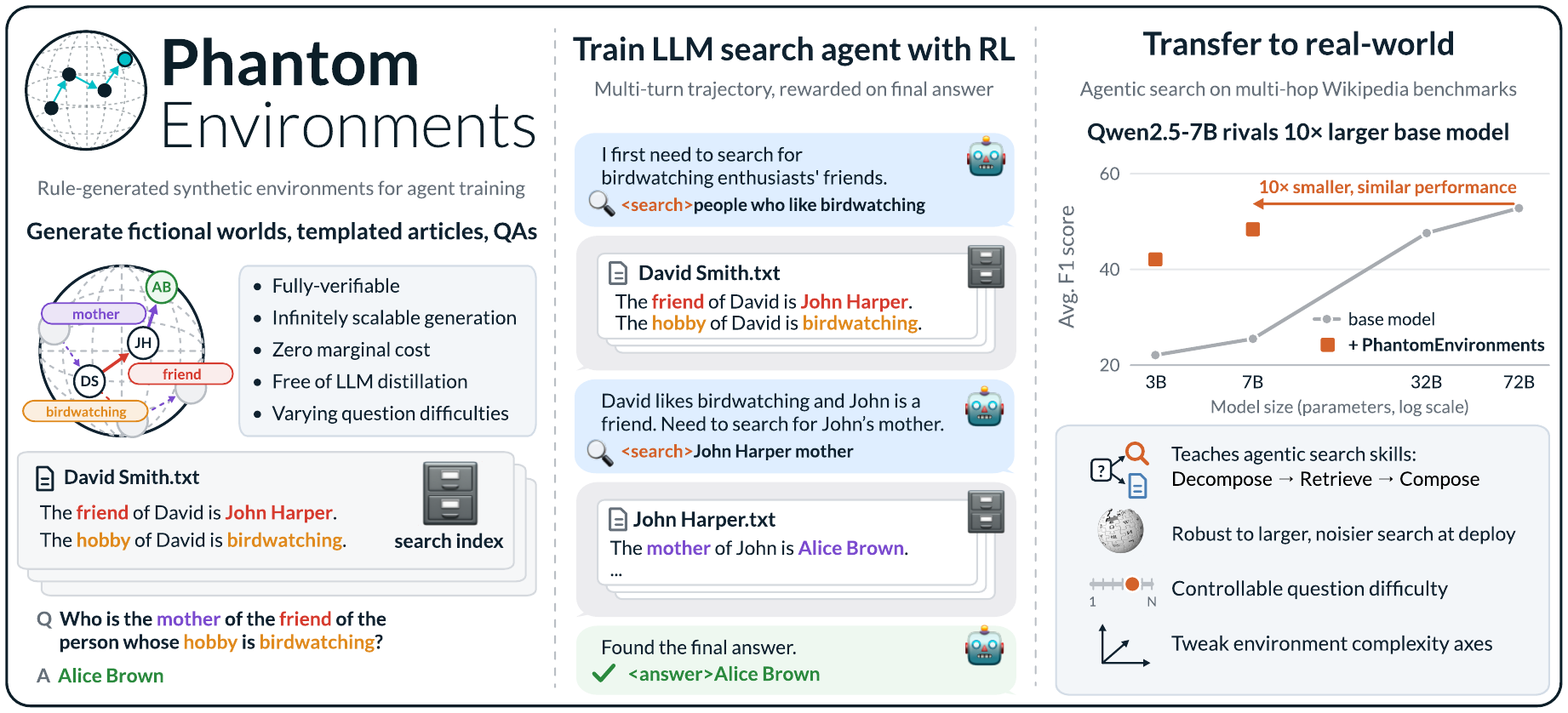}
    \caption{LLM agents trained with RL in \penv transfer to real-world search.}
    \label{fig:motivation}
    \vspace*{-2em}
\end{figure}

Yet, we empirically demonstrate significant and consistent transfer to real-world agentic search.
We RL full fine-tune four LLMs (\qwenthree, \qwenseven~\citep{qwen2.5}, \llamathree~\citep{llama3}, \phimini~\citep{microsoft2025phi}) inside \penv and evaluate on a suite of six real-world multi-hop search benchmarks.
Performance improves over the base model by roughly $1.7\times$ on the Wikipedia-2018-based benchmarks (\hp~\citep{yang2018hotpotqa}, \twowiki~\citep{ho2020constructing}, \msq~\citep{trivedi2022musique}) and $2.2\times$ on the newer-and-harder ones (\synthrm, \synthsm~\citep{gu2025synthworlds} and \frames~\citep{krishna2024frames}).
We also find that trained Qwen models learn to allocate their search budget roughly linearly with question difficulty---an emergent \textit{search scaling} behavior.
To our knowledge, we are the first to show that off-the-shelf LLMs fine-tuned with RL on such rule-generated synthetic environments can create LLM search agents performant in the real-world.

Transfer alone, however, does not tell us how fictional worlds compare to real training data.
Head-to-head against \searchrone-style training in a real Wikipedia-2018 environment, the real-world environment (expectedly) wins on the in-domain Wikipedia-2018 benchmarks.
On benchmarks built from more recent Wikipedia snapshots, however, its advantage narrows for every model and disappears for Qwen models.
The cleanest evidence comes from the SynthWorlds real-vs-synthetic-mirror pair: training with real-world data preserves and even widens the gap between \synthrm and \synthsm evaluation, while training in \penv closes it.
The real environment lets the agent shortcut through memorized Wikipedia facts, whereas the fictional one denies that route and forces models to learn and apply the fundamental search skill.

As generation is rule-based, we can also vary the question distribution to isolate which axes of environment complexity drive transfer.
We design three training environments: linear hop questions, hops compared by attribute, and hops constrained by attribute filters.
We observe that linear hops carry most of the transfer, while comparison questions specifically improve matching real-world comparison questions.
We observe larger gains where the base model is weakest, suggesting a modular path for addressing model capability gaps by composing environment complexities.
Constraint questions, counterintuitively, hurt: agents learn to retrieve gold documents with a single verbatim query that exploits rare attributes.
This causes the models to skip question decomposition entirely, a necessary skill for real-world agentic search.
Environment complexity axes thus do not blindly compose---they help when they match real-world question types and can backfire when they reward shortcuts.

Together, these results position rule-generated synthetic environments as a new source of agent training data, complementary to human-curated and LLM-generated ones.
Each brings something the others cannot: human-curated environments contribute linguistic richness and in-domain evaluation coverage, LLM-synthesized environments enable targeted task design at lower curation effort.
Rule-generated environments add what neither can: training signal for RL that is zero-marginal-cost, exactly verifiable, and never goes stale.
The promise comes with limitations---they lag on in-domain benchmarks where knowledge memorization helps, and must be designed to avoid reward shortcuts.
The core finding stands: environments built from rules alone train capable LLM search agents.

%% file: sections/04_relatedwork.tex
\section{Related Work}
\label{sec:related_work}
\vspace{-0.5em}

\textbf{Reinforcement Learning for LLM Search Agents.}
Following the success of outcome-based RL for reasoning in LLMs~\citep{guo2025deepseekr1}, a line of work has trained LLM search agents to interleave retrieval with chain-of-thought.
Search-R1~\citep{jin2025searchr1} pairs a Wikipedia index with exact-match rewards; follow-ups vary the optimization but not the data source, exploring multi-stage training~\citep{song2025r1searcher}, tighter reasoning--retrieval coupling~\citep{chen2025research}, live web environments~\citep{zheng2025deepresearcher}, step-level reward shaping~\citep{wang2025stepsearch}, and modular designs~\citep{jiang2025s3}.
In this work, we compare the two extremes of synthetic-real environment spectrum: our rule-generated synthetic environments vs the real-world \nqhotpot (multi-hop questions referencing the Wikipedia 2018 corpus~\citep{kwiatkowski-etal-2019-naturalqa,yang2018hotpotqa}).
We treat the environment as the design surface and ask which axes of question complexity drive transfer.

\begin{table}[b]
\centering
\setlength{\tabcolsep}{2pt}
\small
\begin{tabular}{cllc}
\toprule
 & Data source & LLM synthesizes & Verifiability \\
\midrule 
Search-R1 \& followups~\citep{jin2025searchr1} & Human-curated & --- & Human \\
ZeroSearch \citep{sun2025zerosearch}  & LLM-generated   & Search engine responses        & LLM-judged \\
ASearcher \citep{gao2025asearcher}    & LLM-generated   & QA pairs + trajectories        & LLM-judged \\
WebDancer \citep{wu2026webdancer}     & LLM-generated   & Browsing trajectories          & LLM + rejection \\
WebSailor-V2 \citep{li2026websailor}     & LLM-generated   & QA pairs on Wiki seeds & LLM + rejection \\
SWiRL \citep{goldie2025swirl}         & LLM-generated   & Multi-step trajectories        & LLM-judged \\
KARL \citep{databricks2026karl}       & LLM-generated   & Enterprise QA + corpus         & LLM-judged \\
Endless Terminals \citep{gandhi2026endless} & LLM-generated & Terminal task scaffolds   & Execution \\
RandomWorld \citep{sullivan2025randomworld} & LLM-generated & Tool-use instructions & Execution \\
\midrule
\textbf{PhantomEnvironments}            & \textbf{Rule-generated} & ---  & \textbf{Prolog-based, exact} \\
\bottomrule
\addlinespace[2pt]
\end{tabular}
\caption{Data and environments for agent training.
Prior work uses real or LLM-generated data; ours is the first rule-generated, LLM-free pipeline, with zero-marginal-cost and exact verifiability.}
\label{tab:synthetic-data-taxonomy}
\vspace*{-1em}
\end{table}

\textbf{Synthetic Environment Design for Agent Training.}
To reduce the time and cost of hand-curating RL training environments at scale, recent work uses LLMs to synthesize training data: either by simulating retrieval responses~\citep{sun2025zerosearch}, by generating QA pairs and multi-step trajectories~\citep{gao2025asearcher,wu2026webdancer,goldie2025swirl,databricks2026karl,li2026websailor,lu2025deepdive}, or by creating full task harnesses for agents~\citep{gandhi2026endless}.
While LLMs can generate synthetic training data at scale, they introduce hallucination risk and potential benchmark contamination, from real-world knowledge memorized during LLM training.
Creating every training data point incurs non-trivial API cost.
Moreover, LLM-generated data is subject to the LLM's capability ceiling.

A parallel pre-LLM tradition generates training environments procedurally through rules, entirely free of LLMs.
BabyAI~\citep{chevalier2019babyai} established the rule-generated paradigm in gridworlds; RandomWorld~\citep{sullivan2025randomworld} more recently applies type-guided sampling to synthesize tool-use API call sequences---though it still relies on an LLM to populate environment values and instructions.
Both show transfer to real-world benchmarks.
On the evaluation front, PhantomWiki~\citep{gong2025phantomwiki}, SynthWorlds~\citep{gu2025synthworlds}, \gsminf~\citep{zhou2025gsminf} use rule-generated fictional worlds to measure LLM reasoning separately from memorized knowledge.
Closest to our setting, \citet{kabra2026learning} and \citet{stojanovski2025rgym} train on rule-generated synthetic data and show transfer to real-world reasoning.
However, they are limited to the much simpler \emph{in-context reasoning} setting, where \textit{all} relevant documents are supplied in the prompt, and LLMs need only compose knowledge~\citep{kabra2026learning}.
The \textit{agentic reasoning} setting we study is substantially harder: the LLM receives \textit{only} the question and no documents upfront.
It must (1) discover the relevant documents by formulating queries, (2) recover from irrelevant retrievals from the environment, and finally (3) compose knowledge across many environment interactions.
We build on PhantomWiki to an interactive multi-turn RL environment, with no humans or LLMs anywhere in the environment generation pipeline, and exact Prolog-grounded verifiability (\Cref{tab:synthetic-data-taxonomy})~\citep{sterling1994prolog}.
These \penv are thus zero-marginal-cost for generating data.
Moreover, their complexity design is controllable rather than being set by an LLM's capabilities.

\textbf{Multi-Hop Reasoning and Retrieval.}
Multi-hop question answering, requiring evidence chains across multiple documents, has been the dominant testbed for retrieval-augmented reasoning.
A series of increasingly controlled benchmarks---\hp~\citep{yang2018hotpotqa}, \twowiki~\citep{ho2020constructing}, and \msq~\citep{trivedi2022musique}---has driven progress on this task, each tightening controls against reasoning shortcuts.
SynthWorlds~\citep{gu2025synthworlds} and \frames~\citep{krishna2024frames} pose harder variants requiring integration across more sources under complex constraints.
On the method side, IRCoT~\citep{trivedi2023ircot} showed that interleaving retrieval with reasoning beats single-shot retrieval, with later work regulating retrieval through reflection tokens~\citep{asai2023self} or graph-organized corpora~\citep{gutierrez2024hipporag}.
ReAct introduced the \textit{agentic search} setting: agents reason to retrieve information with search queries to a search engine requirement~\citep{yao2022react}.
These methods collectively underscore that effective multi-hop search requires tight coupling between retrieval and reasoning at each step---precisely the skill our synthetic environments are designed to elicit, without relying on any real-world factual content.

%% file: sections/02_methods.tex
\section{Experimental Setup}
\label{sec:experiments}

\subsection{\penv for Agent Training}
Training search agents with RL requires environments that provide verifiable rewards and support long-horizon trajectories.
Rather than relying on human- or LLM-curated data, we build on \pw~\citep{gong2025phantomwiki}, a rule-generated synthetic dataset consisting of multi-hop question-answer pairs about fictional people---these worlds contain no real-world facts.
Fresh synthetic datasets can be generated with varying sizes, hop count, number of questions, and question difficulty.

In these fictional worlds, people are connected through family and friendship relationships; articles about people are generated with programmed templates.
Questions are generated with context-free grammars, and every question-answer pair is fully verifiable by construction.
Concretely, we sample relational chains of up to 7 linear hops over the underlying graph from the grammar (\textit{the sister of the friend of the parent of Alice}).
Each sampled question compiles to a parallel Prolog query that returns ground-truth answers; a question can have multiple answers.
See \Cref{app:synthetic_environment_generation} for details.

We collect all templated articles of individuals in a fictional world into a search index, turning \pw into a search environment where documents can be retrieved by querying. 
The retrieval interface matches the real-world search setting; only the underlying documents are fictional and rule-generated.
This enables an agent to learn---in fictional worlds---how to issue targeted queries and compose knowledge across multiple turns.

\subsection{Training LLMs as Search Agents}
\label{sec:training_llms_as_search_agents}

We follow the RL fine-tuning recipe of \searchrone~\citep{jin2025searchr1} to train models for agentic search, instructing them to use XML tags \searchtags for queries and \answertags for final answers, with retrieved results appended in \infotags. Trajectories terminate when the model outputs a final answer, reaches the max turn limit, or fails to generate these tags.
All details are in \Cref{app:implementation_details}).

\textbf{GRPO and reward design.}
We use the GRPO algorithm where the model generates multiple independent trajectories for a question and is rewarded only on the final answer~\citep{shao2024grpo}.
Following standard practice in RL fine-tuning, we mask out environment's outputs within \infotags when calculating the GRPO objective to update the model's weights~\citep{jin2025searchr1}.
We train on $\approx$55K questions in our synthetic environments.
To compare with real-world training data, we randomly subsample 55K questions (same data budget) from the NaturalQuestions-\hp (\nqhotpot) corpus released with \searchrone~\citep{jin2025searchr1}.
A question in \penv can have multiple ground-truth answers and an LLM can output multiple predictions separated by commas, so we use the F1 score for reward, after SQuAD-style text normalization~\citep{rajpurkar2016squad}.
When questions have 1 final answer, as is the case for \nqhotpot questions, the F1 score reward is the same as Exact Match reward.

\textbf{Training details.}
We fix the training setup so that any difference in transfer is attributable to only the training environment.
To encode documents of each environment and search queries, we use \texttt{intfloat/e5-base-v2} model~\citep{wang2022e5} for dense retriever with a flat FAISS index, and fetch top-3 documents per query~\citep{johnson2019faiss}.
We RL full fine-tune four LLMs of different families and sizes: \qwenthree, \qwenseven, \llamathree, and \phimini~\citep{qwen2.5,llama3,microsoft2025phi}.
We train every setting for 1 epoch, with 2 independent training seeds, in our compute budget of 2 B200s over 2 days.
Throughout evaluation results in \Cref{sec:results} we compute mean and standard errors on both training seeds.
We use the open-source SkyRL library~\citep{cao2025skyrl}.

\subsection{Evaluation Benchmarks}
\label{sec:eval_benchmarks}

We evaluate across Wikipedia-based benchmarks capturing different facets of multi-hop search: standard 2--4 multi-hop datasets (\hp, \twowiki, \msq), newer 2--6 hop benchmarks to disentangle memorization and reasoning (\synthrm, \synthsm), and a harder benchmark requiring longer reasoning chains of 2--15 hops (\frames).
Wikipedia reference articles of these benchmarks span across LLM knowledge cutoff dates.
For \hp, \twowiki, \msq, and \synthrm, we use 500 evaluation questions from \citep{kabra2026learning}; for \synthsm, we use the 500 synthetic-mirror counterparts of \synthrm; and for \frames, all 824 questions with June 2023 Wikipedia articles using the specified URLs.
We additionally use 500 questions of \cofca~\citep{wu2024cofca} benchmark for our ablation on environment complexities (\Cref{sec:env_design}).
Details in \Cref{app:evaluation_benchmarks}.

\textbf{Evaluation setup.}
Each benchmark contributes its own retrieval corpus---built from the gold reference and distractor paragraphs accompanying each question.
For \frames, we use section-level chunks of the Wikipedia reference articles.
We report token-level F1 score for these real-world benchmarks, computed with the canonical SQuAD-based scorer of each benchmark (\hp-style yes/no/noanswer guard for \hp, \twowiki, and \cofca; \msq-style F1 scoring otherwise).
At evaluation time we use a retriever setup of the maximum 32768 context limit, up to 20 turns, and the Qwen3-Embedding-4B~\citep{qwen3embedding} dense retriever---this setup is superior to our training setup.
Following prior work~\citep{jin2025empirical}, we find that this superior evaluation configuration yields better reported benchmark performance for all LLMs, trained or otherwise.
See \Cref{tab:retriever_qwen3b,tab:retriever_qwen7b} in \Cref{app:retrieval_server_details} for ablation results on retriever choice.

%% file: sections/03_results.tex
\section{Results}
\label{sec:results}

\subsection{Performance Transfer from \penvshort to Real-World Evaluation}

\input{images_arxiv/f1_table}

\begin{figure}[!h]
    \vspace{-0.5em}
    \centering
    \includegraphics[width=\linewidth]{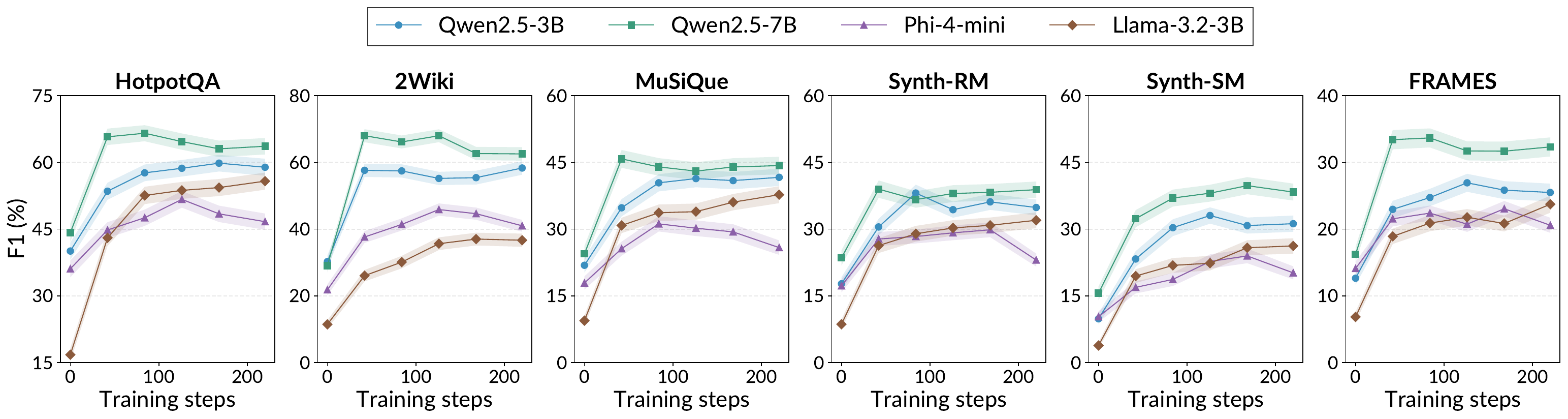}
    \caption{%
    \textbf{F1 scores steadily improve as training progresses.}
    We evaluate intermediate checkpoints of \penv training runs, and observe steady performance improvements across the board: a sharp increase initially then a steady growth.
    While some runs saturate, we generally do not see drastic overfitting or collapse to the rule-generated templates of \penv.
    We report mean $\pm$ standard error as the solid line and shaded region.}
    \label{fig:f1_vs_training_steps}
    \vspace{-0.5em}
\end{figure}

\Cref{tab:f1} reports the F1 scores on real-world multi-hop agentic search benchmarks before and after RL fine-tuning in \penv.
Training on synthetic environments consistently yields significant gains across all six benchmarks and four LLMs (\qwenthree, \qwenseven, \llamathree, \phimini).
Relative to the base model, \llamathree improves $7.1\times$ (from $3.8$ F1 score to $27.0$ on \synthsm), and at the minimum by $3.2\times$ for other benchmarks.
Both \qwenthree and the larger \qwenseven improve on average $1.9\times$ after RL fine-tuning in \penv, so the performance transfer holds when scaling LLM size. 
Overall, the gains average $1.7\times$ F1 on older benchmarks pre-LLM cutoff dates (\hp, \twowiki, \msq) and $2.2\times$ on more recent and challenging benchmarks (\synthrm, \synthsm, and \frames): these LLM agents can search in fresh and more difficult real-world settings.
Over the course of synthetic training, models do not overfit to the templated questions and fictional universes, and continue to improve on benchmarks (\Cref{fig:f1_vs_training_steps}).
Every training question is only seen once, so training steps is a proxy for training samples.
Hence, this shows the benefit of \textit{data scaling} in LLM agent training.

\textbf{Robustness to larger-scale deployment with noisy search.}
When agents are deployed, they encounter larger search environments where corpus documents can overlap and interfere.
In practice, deployment environments can return noisier search results to agent queries than the agents were trained to handle.
This demands for training environments that create agents robust to noisy search and ready for larger-scale deployment---do our \penv meet this bar?
\input{images_arxiv/pooled_table_qwen7b}
We construct this test scenario by gathering the documents of all six benchmarks into one large \textit{pooled search index} that simulates noisy search.
The pooled corpus is, at the minimum, $1.5\times$ the size of \frames' corpus alone, and up to $44\times$ for SynthWorlds pair.
Moreover, it contains interfering and near-duplicate documents: (1) Wikipedia passages of \hp and \msq overlap with long \frames articles, and (2) \synthrm documents compete with \synthsm as they are identical except for real vs fictional facts~\citep{gu2025synthworlds}.
In \Cref{tab:pooled_qwen7b} we ablate agent performance on the choice of retrieval corpus, either benchmark's own or pooled corpus.
Every agent degrades, but only \textit{slightly}: the \qwenseven base worsens by $1.6$ average F1 score, and the synthetic-trained ones by $1.9$ points (see \Cref{tab:pooled_qwen3b,tab:pooled_llama3b} for more results).
Indeed, agents trained in simplified fictional worlds are robust to large-scale deployments.

\begin{figure}[!h]
    \vspace{-0.5em}
    \centering
    \includegraphics[width=\linewidth]{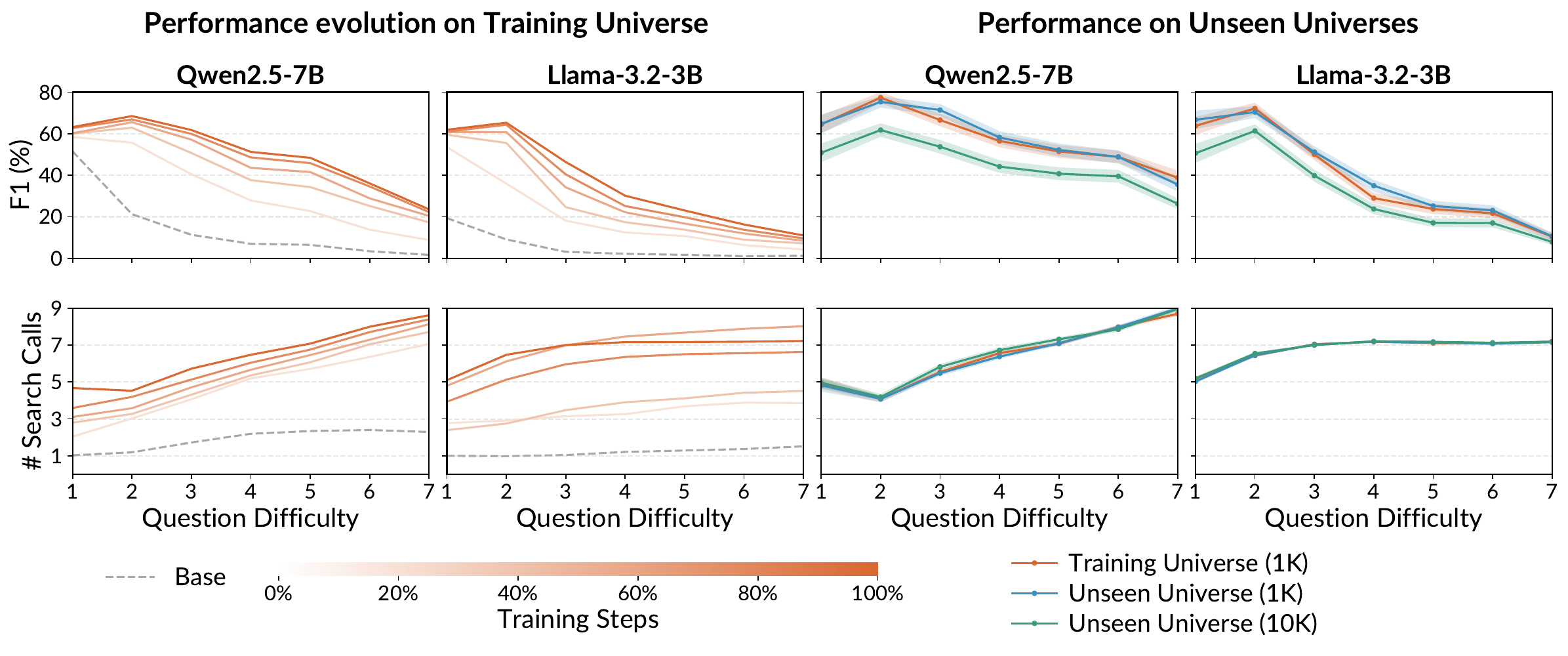}
    \caption{%
    \textbf{F1 scores and number of search calls as a function of question difficulty (hops).}
    \textit{(Left two)} We evaluate intermediate training checkpoints on validation questions from the training universe.
    As fine-tuning progresses, F1 increases across all difficulty levels for both LLMs (darker lines are higher in scores).
    In the lower panels, we plot the number of search calls as a function of question difficulty.
    We observe an emergent ``search scaling'' property in Qwen2.5 models: number of searches scales linearly with environment's question difficulty.
    \llamathree shows partial search scaling, increasing search calls initially and plateauing.
    \textit{(Right two)} We evaluate final checkpoints on an unseen universe of the same size as training (Unseen~1K) and a $10\times$ larger universe (Unseen~10K).
    F1 scores and search scaling behavior in Unseen universes parallel the Training universe, indicating agents acquired generalizable agentic search skill rather than memorizing facts.}
    \vspace{-0.5em}
    \label{fig:pw_generalization_vs_difficulty_qwen7b_llama}
\end{figure}

\textbf{Training evolution and emergent search scaling behavior.}
In addition to real-world transfer, \penv allow for fine-grained analysis as they are fully verifiable and deterministically generated.
At environment generation time, question difficulties---the number of fictional Wikipedia documents to navigate---is known.
Using this feature, we analyze agents' performance evolution as RL fine-tuning progresses.
In  \Cref{fig:pw_generalization_vs_difficulty_qwen7b_llama} (left) we find that F1 scores of \qwenseven and \llamathree improve on questions of all difficulties as training progresses (\Cref{fig:pw_generalization_vs_difficulty_qwen3b_phi} visualizes evolution of \qwenthree and \phimini).
However, a key behavioral difference emerges between LLM families: Qwen2.5 ones learn to automatically issue search calls proportional to question difficulty, exhibiting a clear linear relationship.
We call this desirable property ``search scaling'', emergent in Qwen2.5 LLMs from environment interaction alone.
\llamathree shows partial search scaling: number of searches grows linearly with question difficulty, then plateaus.
On the other hand, \phimini in \Cref{fig:pw_generalization_vs_difficulty_qwen3b_phi} does not display any scaling.
This suggests that this property is model-dependent and emergent only for sufficiently capable LLMs.
This finding aligns with prior work on how model baseline capabilities influence benefits from further RL fine-tuning~\citep{gandhi2025cognitive, yue2025does}.

\textbf{Generalization in unseen environments.}
One concern with training on a fixed fictional universe is that the LLM could simply memorize the knowledge graph's facts and retrieve answers by pattern matching.
To test fact memorization, in \Cref{fig:pw_generalization_vs_difficulty_qwen7b_llama} (right) we evaluate on a freshly-generated unseen environment of the same size (Unseen~1K) and a $10\times$ larger universe (Unseen~10K).
These evaluation environments share no factual knowledge overlap with \textit{any} training data, seen either during pretraining or RL fine-tuning.
The trained agents match their in-domain F1 score performance at every difficulty level, when the universe size is fixed at 1K.
The trend follows at the larger universe of 10K as well, where the performance is slightly worse than Unseen~1K due to the more noisy search environment.
Results for \qwenthree and \phimini in \Cref{fig:pw_generalization_vs_difficulty_qwen3b_phi}.
Robust performance to unseen environments confirms that LLM search agents have learned the generalizable agentic search skill.

\subsection{Comparison with Real-World Training Environments}

So far we have showed that fictional worlds can teach agentic search.
A natural next question is how their transfer compares to training on real-world data.
We find that real-world training environments---with its linguistic richness and grounded factual knowledge---outperform synthetic when evaluation benchmarks contain in-domain facts to training, e.g. when Wikipedia cutoffs overlap.
When evaluation becomes out-of-domain, however, \penv outperform real-world.

\begin{figure}[!h]
    \vspace{-0.5em}
    \centering
    \includegraphics[width=\linewidth]{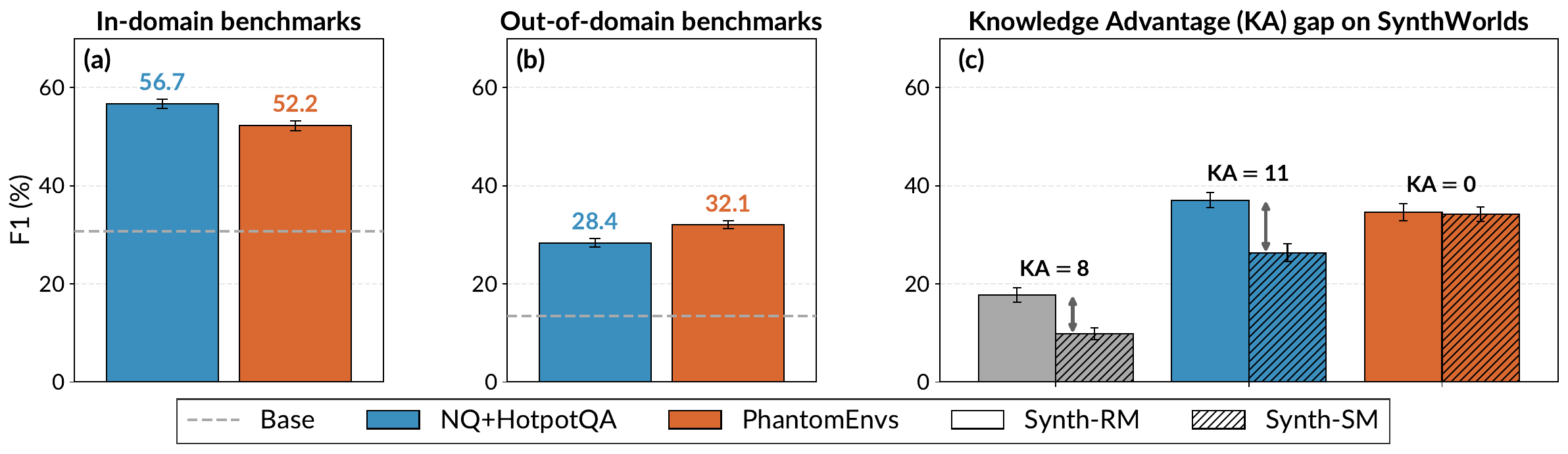}
    \caption{%
    \textbf{(a)}
    Training \qwenthree on real-world \nqhotpot data outperforms \penv on benchmarks in-domain to \nqhotpot (\hp, \twowiki, \msq).
    \textbf{(b)}
    Whereas \penvshort is better than real-world training on newer and harder out-of-domain benchmarks (\synthrm, \synthsm, \frames).
    \textbf{(c)}
    \penvshort teach agents to be equally performant on SynthWorlds pair: F1 scores are similar for RM and SM versions (KA $= 0$), but KA $> 0$ for the base model and \nqhotpot training.
    See text for details.
    }
    \vspace{-0.5em}
    \label{fig:f1_transfer_performance_real_vs_pw_qwen3b}
\end{figure}

\textbf{Real-world environments are superior when in-domain to evaluation.}
In \Cref{fig:f1_transfer_performance_real_vs_pw_qwen3b} we compare \penv training against a real-world training environment: \nqhotpot questions from the 2018 Wikidump released with Search-R1~\citep{jin2025searchr1}.
Three benchmarks (\hp, \twowiki, \msq) share the same Wikipedia~2018 corpus, and overlap in factual knowledge to \nqhotpot training.
Expectedly, when training and evaluation are in-domain, real-world training data outperforms synthetic for all LLMs: \qwenthree in \Cref{fig:f1_transfer_performance_real_vs_pw_qwen3b}~(a) gets an average $56.7\%$ F1 score vs $52.2\%$ respectively (see \Cref{fig:f1_transfer_performance_real_vs_pw_qwen7b,fig:f1_transfer_performance_real_vs_pw_llama3,fig:f1_transfer_performance_real_vs_pw_phi} for other LLM results).

\textbf{On newer out-of-domain benchmarks, \penv transfers better.}
The story changes on \synthrm, \synthsm, and \frames benchmarks that are sourced from Wikipedia post 2023.
\nqhotpot data now has limited factual knowledge overlap, so evaluation is out-of-domain in factual knowledge.
Here \penv outperform real-world data: \qwenthree in \Cref{fig:f1_transfer_performance_real_vs_pw_qwen3b}~(b) with \penvshort gets an average $32.1\%$ F1 score vs $28.4\%$.
This suggests that \penv yield LLM agents robust to distribution shift.
We further corroborate this hypothesis with the SynthWorlds pair: \synthrm (real Wikidata entities) and \synthsm (the same graph re-instantiated with synthetic entities)~\citep{gu2025synthworlds}.
This pair calculates a Knowledge Advantage (KA) gap, which is positive when LLMs utilize memorized knowledge from Wikipedia.
\Cref{fig:f1_transfer_performance_real_vs_pw_qwen3b}~(c) shows KA gaps for base and trained models.
Base models already exhibit positive KA gap---\qwenthree has $8\%$, and that widens to $11\%$ after real-world \nqhotpot training.
Remarkably, our \penv \textit{fully close} the KA gap to $0\%$.
For other LLMs (\Cref{fig:f1_transfer_performance_real_vs_pw_qwen7b,fig:f1_transfer_performance_real_vs_pw_llama3,fig:f1_transfer_performance_real_vs_pw_phi}), synthetic can reduce the KA gap when real-world data amplifies it.

This contrast shows the need for training environments that are robust when evaluation targets fall beyond LLM training cutoffs---and our \penv meet this need.
Both human-curated and LLM-generated data are anchored to a timed snapshot of the world, so benchmarks built on later snapshots eventually fall outside their coverage.
Rule-generated synthetic environments are decoupled from this temporal drift, \textit{timeless} in a way, and robust to knowledge memorization.
In new evaluation environments where such shortcuts through memorization are absent, \penv generalize better than real-world training.

%% file: images_arxiv/f1_table.tex
\begin{table}[!h]
\small
\centering
\setlength{\tabcolsep}{4pt}
\begin{tabular}{lcccccc}
\toprule
Model & HotpotQA & 2Wiki & MuSiQue & Synth-RM & Synth-SM & FRAMES \\
\midrule
Qwen2.5-3B & $40.1 \pm 1.9$ & $30.3 \pm 1.9$ & $21.9 \pm 1.6$ & $17.7 \pm 1.5$ & $9.9 \pm 1.2$ & $12.6 \pm 1.0$ \\
\addlinespace[2pt]
\rowcolor{lightorange} \quad + PhantomEnvs & $60.9 \pm 1.5$ & $56.4 \pm 1.5$ & $39.4 \pm 2.1$ & $34.6 \pm 1.8$ & $34.2 \pm 1.4$ & $27.4 \pm 1.0$ \\
\addlinespace[2pt]
Qwen2.5-7B & $44.2 \pm 2.0$ & $29.1 \pm 1.9$ & $24.5 \pm 1.8$ & $23.6 \pm 1.7$ & $15.6 \pm 1.5$ & $16.2 \pm 1.1$ \\
\addlinespace[2pt]
\rowcolor{lightorange} \quad + PhantomEnvs & $64.1 \pm 2.3$ & $66.7 \pm 2.2$ & $44.7 \pm 2.8$ & $39.1 \pm 1.8$ & $40.6 \pm 1.8$ & $35.4 \pm 1.1$ \\
\addlinespace[2pt]
Llama-3.2-3B & $16.8 \pm 1.5$ & $11.5 \pm 1.3$ & $9.4 \pm 1.1$ & $8.6 \pm 1.1$ & $3.8 \pm 0.7$ & $6.8 \pm 0.8$ \\
\addlinespace[2pt]
\rowcolor{lightorange} \quad + PhantomEnvs & $55.8 \pm 1.6$ & $37.5 \pm 1.5$ & $35.9 \pm 1.8$ & $30.8 \pm 2.1$ & $27.0 \pm 1.3$ & $23.4 \pm 1.1$ \\
\addlinespace[2pt]
Phi-4-mini & $36.1 \pm 1.9$ & $21.8 \pm 1.6$ & $17.9 \pm 1.5$ & $17.3 \pm 1.5$ & $10.3 \pm 1.2$ & $14.1 \pm 1.0$ \\
\addlinespace[2pt]
\rowcolor{lightorange} \quad + PhantomEnvs & $47.6 \pm 1.8$ & $42.3 \pm 1.7$ & $27.2 \pm 1.6$ & $26.7 \pm 1.5$ & $18.3 \pm 1.4$ & $21.6 \pm 1.1$ \\
\bottomrule
\addlinespace[2pt]
\end{tabular}
\caption{%
    \textbf{F1 scores on real-world agentic search benchmarks after RL fine-tuning in \penv.}
    Synthetic training significantly improves performance of all LLM families and sizes.
    Improvements are largest on the newer and harder benchmarks---\synthrm, \synthsm, and \frames---\llamathree improves by up to $7.1\times$ on \synthsm.
    We report mean $\pm$ standard error over the test sets and two training seeds.
    }
\label{tab:f1}
\vspace*{-0.5em}
\end{table}

%% file: images_arxiv/pooled_table_qwen7b.tex
\begin{wraptable}{r}{0.47\linewidth}
\centering
\small
\setlength{\tabcolsep}{2pt}
\begin{tabular}{lcc}
\toprule
Corpus & Qwen2.5-7B & + PhantomEnvs \\
\midrule
Benchmark's own & $25.5 \pm 0.7$ & $48.4 \pm 0.8$ \\
\rowcolor{lightorange} All pooled & $23.9 \pm 0.7$ & $46.5 \pm 0.8$ \\
\bottomrule
\addlinespace[2pt]
\end{tabular}
\caption{%
    \textbf{Average F1 scores of \qwenseven, ablating choice of search corpus.}
    Each benchmark's questions are answered either against that benchmark's own corpus, or against a single pooled corpus.
    Expectedly, agents perform worse against the pooled corpus but the drop is slight, with or without training.
    }
    \vspace*{-1em}
\label{tab:pooled_qwen7b}
\end{wraptable}

%% file: sections/05_analysis.tex
\section{Which Environment Complexity Axes Drive Transfer in Agents?}
\label{sec:env_design}

Real-world questions span a spectrum of complexity, such as requiring multiple hops over entities, comparing and combining parallel hop chains, and filtering candidates by constraints.
We ablate our synthetic environments along three orthogonal axes of complexity: \emph{linear hops} over entities, \emph{comparisons} of attributes, and \emph{constraints} that filter candidates.
Holding the corpus fixed, we generate these synthetic environments by extending \penv generation code, context-free-grammars, and Prolog queries.
We then fine-tune \qwenthree and \llamathree.

The original \textbf{\hops} environment has questions of up to 7 linear hops from an anchor person (\textit{Who is the friend of parent of Alice}) or a tail attribute (\textit{whose hobby is birdwatching}).
The new \textbf{\hopscomp} environment pairs two such chains and adds a comparison predicate to the front (\textit{Who is older, friend of parent of Alice, or sister of Bob?}).
We balance the training mixture to 50--50 between pure-hop and comparison questions.
Finally, the new \textbf{\hopscons} environment inserts up to three attribute filters at arbitrary positions along the chain (\textit{Who is the friend, whose hobby is reading, of parent of Alice?}), with the same 50--50 balance between pure-hop (0--1 filters) and multi-constraint (2--3 filters) questions.
Full specifications are in \Cref{app:synthetic_environment_generation}.

\begin{wrapfigure}{r}{0.5\textwidth}
    \centering
    \includegraphics[width=0.5\textwidth]{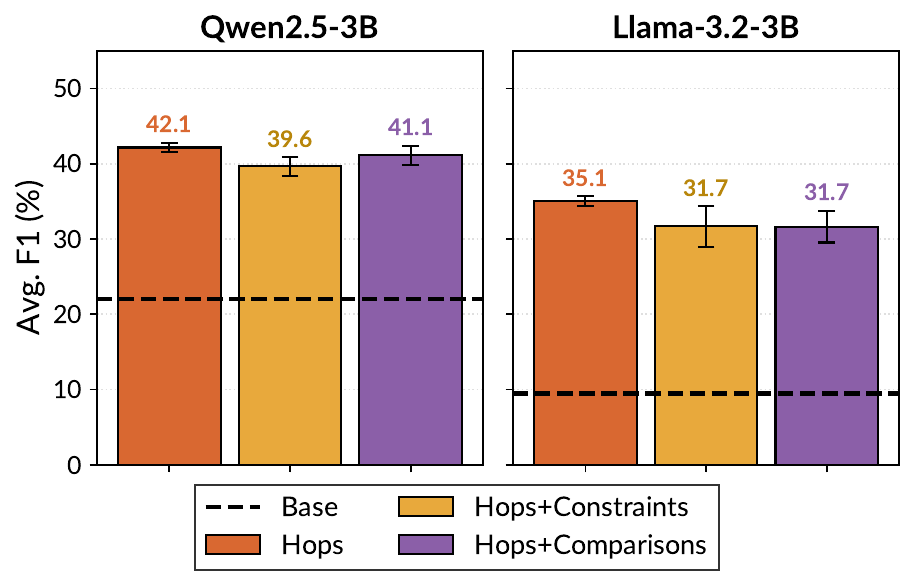}
    \caption{%
    \textbf{Ablation results on synthetic environment complexity axes.}
    We find that \hops is the strongest axis that drives real-world transfer.
    }
    \label{fig:f1_transfer_performance_pw_vs_variants}
    \vspace{-1em}
\end{wrapfigure}
\textbf{Not every complexity axis equally drives transfer.}
One might expect that adding complexity axes only adds to performance transfer.
However, \Cref{fig:f1_transfer_performance_pw_vs_variants} shows otherwise---\hops is at least as good as \hopscomp and \hopscons, so linear hops is the dominant complexity axis for performance transfer.

\begin{figure}[!h]
    \vspace{-1em}
    \centering
    \includegraphics[width=0.8\linewidth]{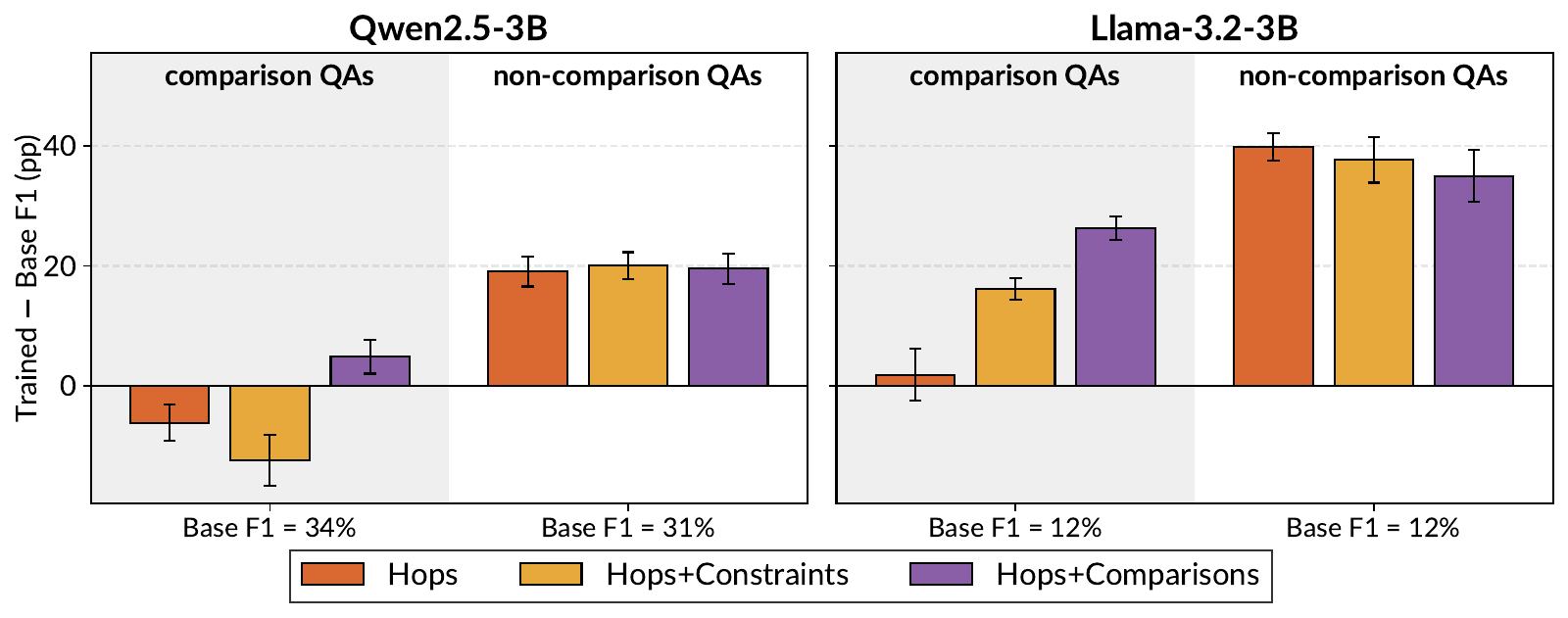}
    \caption{%
    \textbf{F1 score deltas per question type of \cofca benchmark after fine-tuning on environment variants.}
    We plot F1 score deltas on comparison and non-comparison questions: we calculate F1 score difference per question, average deltas for each category, and report the mean $\pm$ standard error (paired) in percentage points.
    \hopscomp environment improves performance on evaluation comparison questions, but linear hops remains dominant otherwise.
    }
    \label{fig:f1_category_delta_pw_variants_cofca}
    \vspace{-1.5em}
\end{figure}

\textbf{Where exactly do the complexity axes help or hurt?}
Aggregate F1 scores obscure behavioral changes in agents introduced by environment variants. 
In \Cref{fig:f1_category_delta_pw_variants_cofca,fig:f1_category_delta_pw_variants_twowiki} we decompose performance by question category on \cofca and \twowiki benchmarks, which contain category labels for questions (see  \Cref{app:comparison_questions} for details on category labels).
Both have comparison questions, which our respective environments \hopscomp and \hopscons directly target.
We note three observations.
First, comparison training questions targets the matching real-world category: \hopscomp significantly outperforms the \hopscons and just \hops.
The gain from \hopscomp is most pronounced for \llamathree, which has a low base F1 score for \cofca comparison questions.
So environment complexity axes can improve base models on question types on which they are particularly weak.
Second, linear hops remain the dominant axis of complexity for transfer on non-comparison categories.
Third, we find shortcutting behavior with \hopscons---agents issue verbatim search queries rather than decomposing---and that actively hurts transfer (\Cref{app:hopscons_shortcut_trajectories}).

%% file: sections/05_conclusion.tex
\section{Conclusion}
\label{sec:conclusion}

We show that LLM search agents trained in \penv, with no humans or LLMs in the data pipeline, transfer to real-world multi-hop search.
Real training data wins where memorized knowledge helps, but its advantage narrows on benchmarks built from newer Wikipedia snapshots.
Ablating environment complexity reveals that linear hops drive most of the transfer, comparison questions target their matching real-world category, and constraint questions backfire by rewarding a verbatim-query shortcut.
In this work, we find rule-generated synthetic environments as a zero-marginal-cost and effective source of agent training data.

\textbf{Limitations.}
Our complexity analysis invites further work on environment composition: given a known capability gap, can one read off a recipe of synthetic axes that closes it?
We use synthetic environments as standalone; blending them with real-world data, e.g., as mid-training data, is a natural next step.
Compute budgets limit us from training larger LLMs and evaluating on long-context and open-web benchmarks like BrowseComp-Plus~\citep{chen2025BrowseCompPlus}---all open directions.

%% file: appendix/01_implementation_details.tex
\section{Implementation Details}
\label{app:implementation_details}

\subsection{Instruction Prompt}
\label{app:instruction_prompt}

We use the following chat-style system and user prompts for all training and evaluation.

\begin{verbatim}
system:

You are a helpful and harmless assistant.

user:

Answer the given question using the search tool to retrieve relevant
information. Follow this process:
1. Reason about what you know and what you still need to find out with
the search tool. For complex questions, break them into sub-questions
and search for each part.
2. If you need more information, call the search engine:
<search> query </search>. Write focused, specific queries. If a query
returns unhelpful results, try a different phrasing or a more specific
sub-question.
3. Base your final answer only on information from the search results,
not on prior knowledge.
4. Once you have enough information, provide your answer inside
<answer> and </answer>. For multiple answers, use a comma-separated
list: <answer>Alice, Bob</answer>. Do not include explanations inside
the answer tags.

Question: {question}
\end{verbatim}

\subsection{RL Fine-tuning}
\label{app:rl_finetuning_details}

We optimize the policy with Group Relative Policy Optimization (GRPO)~\citep{shao2024grpo,guo2025deepseekr1}, following \searchrone~\citep{jin2025searchr1} and the SkyRL implementation~\citep{cao2025skyrl}.
For each question we sample $G=8$ rollouts from the current policy and assign each trajectory a reward: F1 score reward between 0 and 1 for training in our rule-generated synthetic environments.
For \nqhotpot training of \searchrone that has only 1 ground-truth answer, F1 score becomes Exact Match reward of 0 or 1.

Advantages in GRPO are normalized per group of rollouts:
\begin{align*}
A_i = \frac{r_i - \mathrm{mean}(\set{r_1, \dots, r_G})}
           {\mathrm{std}(\set{r_1, \dots, r_G})}.
\end{align*}

The policy is updated by maximizing the token-level clipped surrogate
\[
\mathcal{L}(\theta) = \mathbb{E}_{i, t}\!\left[
   \min\!\left(
       \rho_{i,t}(\theta)\, A_i,\;
       \mathrm{clip}\!\left(\rho_{i,t}(\theta),\,
                            1 - \epsilon,\,
                            1 + \epsilon\right) A_i
   \right)
\right],
\quad
\rho_{i,t}(\theta)
  = \frac{\pi_\theta(y_{i,t} \mid x, y_{i,<t})}
         {\pi_{\theta_{\mathrm{old}}}(y_{i,t} \mid x, y_{i,<t})},
\]
where the expectation runs only over tokens generated by the model.
Tokens inside \infotags, which are returned by the search environment rather than produced by the policy, are masked out of both the importance ratio and the loss.

We clip-high the importance ratios at token-level at $2.0$, and remove the KL penalty against the reference policy.

\textbf{Optimization.}
We use AdamW with a learning rate of $1 \times 10^{-6}$, gradient norms clipped at $1.0$, and a linear warmup over the first $10\%$ of training steps.
Each batch contains 256 question prompts with 8 rollouts per prompt.
The maximum prompt length is 8192 tokens, and each trajectory is capped at 10 environment turns with at most 500 generated tokens per turn.
Trajectories get automatically truncated if they exceed, but we don't observe this during training.
The policy is trained in full precision (fp32) using FSDP2 for sharding; rollout generation is handled by vLLM.
We use the following sampling temperature and nucleus probability at both training and evaluation time:
\begin{enumerate}
    \item Qwen models: $1.0$ temperature, $1.0$ top-p.
    \item Llama-3.2-3B-Instruct: $0.6$ temperature, $0.9$ top-p.
    \item Phi-4-mini-instruct: $0.8$ temperature, $0.95$ top-p.
\end{enumerate}

\textbf{Hardware and software.}
Every run uses 2 NVIDIA B200 GPUs and the open-source SkyRL FSDP2 backend at commit \texttt{298bebc}~\citep{cao2025skyrl}.
All training experiments take up to 2 days on this configuration.

\textbf{Real-world training data.}
For the comparison against in-domain real-world training in \Cref{sec:results}, we draw from the \nqhotpot training corpus released with \searchrone~\citep{jin2025searchr1}.
That corpus contains roughly 160K questions; we shuffle and subsample 55K of them so the data budget matches our synthetic-environment runs exactly.

\textbf{Training dynamics and instability}.
For a couple of real-world training data runs, we observed entropy spikes and training degeneration in \phimini and \llamathree models, often $\approx 50\%$ into the run.
This occurred on 1 training seed for \llamathree on synthetic ablation \hopscons as well.
In such cases, we report the prior saved policy checkpoint with the best held-out set performance.
We rule out \penv as the cause, as we observe model degeneracy in real-world training data and one-off training seeds.
Such RL training instability is observed in practice, even at frontier scale: MAI-Thinking-1 frequently observed crashes and restarted with heuristics~\citep{microsoft2026mai}.
Identifying the cause is an interesting open research direction.

\subsection{Retrieval Server}
\label{app:retrieval_server_details}

At every search turn, the LLM policy queries a locally served dense retrieval index built with FAISS~\citep{johnson2019faiss}.
We use exact (flat) search and return the top-3 passages for each query.
Each evaluation benchmark contributes its own retrieval corpus, constructed from the gold and distractor paragraphs accompanying its question set.
The one exception is \frames, for which we download Wikipedia articles at the June 2023 revision cutoff that matches the benchmark's authoring snapshot, and chunk them into section-level passages.

During training, both the synthetic and real-world corpora are encoded with \texttt{intfloat/e5-base-v2} (768 dimensions)~\citep{wang2022e5}.
This is the same encoder used by \searchrone, so the training retriever is fixed across synthetic and real-world environments, ensuring that differences in downstream performance reflects the training data source.

At evaluation time we switch to a stronger encoder, \texttt{Qwen3-Embedding-4B} (2560 dimensions), and apply it uniformly across all evaluation benchmarks.
Prior work has noted that retriever quality at evaluation time is a major confound when comparing LLM agents, and recommend using the best available search retriever index~\citep{jin2025empirical,chen2025BrowseCompPlus}.
Therefore, we standardize evaluation on the high-quality \texttt{Qwen3-Embedding-4B} retriever, and isolate the significance of our results to the training setup alone.
In \Cref{tab:retriever_qwen3b,tab:retriever_qwen7b}, evaluation results are slightly worse---across the board---when we use the training setup during evaluation.

\input{images_arxiv/retriever_table_qwen3b}

\input{images_arxiv/retriever_table_qwen7b}

\subsection{Synthetic Environment Generation}
\label{app:synthetic_environment_generation}
 
Universes and questions are produced by the PhantomWiki
generator~\citep{gong2025phantomwiki} and our extensions for
constraint-bearing and comparison questions. The training universe
contains $1{,}000$ individuals organized into $100$ family trees of up
to $10$ people each, with a maximum tree depth of $10$, using
PhantomWiki's easy-mode relation set (questions contain only friendship or immediate family relationships, no derived relationships like grandfather).
Each individual is rendered as a single templated article in a wiki-style format; these articles form the documents in the retrieval corpus.
 
Three question-generation configurations are used. In all three,
aggregation questions (``how many\ldots'') are filtered out (easy to reward-hack), and gold answers are obtained by executing the Prolog query attached to each template against the universe graph. A single question may therefore admit multiple gold answers.
 
\begin{itemize}[noitemsep]
    \item \textbf{\hops.} Questions are generated at CFG depth $16$, producing chains of up to $7$ relational hops. We obtain roughly 55K training questions and 5K validation questions after filtering out aggregation questions.
 
    \item \textbf{\hopscons.} The same base
        generator with up to $7$ hops and up to $3$ attribute
        constraints per question, drawn from all base question types.
 
    \item \textbf{\hopscomp.} Up to $7$ hops and at
        most $1$ constraint. The question type pool is restricted to
        base questions and age-based comparisons.
\end{itemize}
 
\paragraph{Cell-balanced sampling.}
Training sets are balanced so that no difficulty stratum dominates the
mixture. For \hopscons, we partition questions into cells of a
two-dimensional grid indexed by $(\text{hops}, \text{constraints})$ and
downsample every cell to the size of the smallest cell. This produces a
uniform distribution across the grid and yields the $50/50$ split
between pure-hop and multi-constraint questions reported in
\Cref{sec:env_design}. For \hopscomp, we partition by
$(\text{hops}, \text{kind})$ where kind is one of \{base, comparison\},
determined from the question surface form. Because the zero-hop
comparison cell is small, we apply a soft cap rather than strict
equalization: each cell is capped at four times the size of the smallest
cell. In both regimes, $10\%$ of each balanced cell is held out as a
stratified validation split.
 
\paragraph{Out-of-domain universes.}
The generalization experiment in
\Cref{fig:pw_generalization_vs_difficulty_qwen7b_llama} uses two additional universes generated with different random seeds: an identically sized 1000-person universe (Unseen~1K) and a larger 10000-person universe (Unseen~10K).
Generator configurations and command lines are released with the code.

\subsubsection{An Improved Bidirectional Anchor Sampling}
\label{app:bidirectional_sampling}
 
PhantomWiki's default question sampler grounds relation chains from one
fixed end: it picks a starting entity and walks forward through
relations until the chain is fully instantiated, with the answer always
landing at the final position. This introduces a positional bias. Entities
that sit at the periphery of the knowledge graph (leaves of family trees,
people with few friends) are over-represented as answers, because they are
easy to reach as chain endpoints but rarely serve as starting points.
 
We replace this with a \emph{bidirectional anchor sampling} procedure
that allows the answer to occupy any position in the chain. The idea is
straightforward: instead of committing to a direction before walking, we
first choose an anchor point inside the chain and then extend outward in
both directions.
 
\paragraph{Setup.}
Let $G = (V, E)$ be the universe graph, where each vertex $v \in V$
represents a person and edges carry typed relations from the set
$R$ (e.g., \texttt{parent}, \texttt{sibling}, \texttt{friend}).
Each person also carries a small set of attributes (date of birth,
occupation, hobby). A question template, produced by the CFG, specifies
a sequence of $k$ typed relation slots that must be filled to form a
valid chain $v_0 \xrightarrow{r_1} v_1 \xrightarrow{r_2} \cdots
\xrightarrow{r_k} v_k$. In the original sampler, $v_0$ is always the
answer.
 
\paragraph{Procedure.}
Given a template with $k$ relation slots:
\begin{enumerate}[noitemsep]
    \item \textbf{Choose an anchor.} Select an anchor index
        $a \in \{0, \ldots, k\}$ uniformly at random. Sample a person
        $v_a$ from $V$.
 
    \item \textbf{Walk backward} (from $v_a$ toward slot $0$). For each
        slot $i = a{-}1, a{-}2, \ldots, 0$, query the universe for
        entities $u$ satisfying $r_{i+1}(u, v_{i+1})$ and sample one
        uniformly. This builds the identifying description of the
        anchor: the noun phrase that a reader would follow to locate
        $v_a$ in the corpus.
 
    \item \textbf{Walk forward} (from $v_a$ toward slot $k$). For each
        slot $j = a{+}1, a{+}2, \ldots, k$, query for entities $w$
        satisfying $r_j(v_{j-1}, w)$ and sample one uniformly. The
        entity at the end of this walk, $v_k$ (or $v_0$, depending on
        which end is designated the answer), becomes the gold answer.
 
    \item \textbf{Verify uniqueness.} Run the full Prolog query
        instantiated with the sampled entities. If the query returns
        a unique answer, accept the sample. Otherwise, discard and
        resample from step~1.
\end{enumerate}
 
The anchor strategy can be set to \emph{uniform} (the default described
above), \emph{balanced} (cycling through start, middle, and end
positions equally), or fixed at a specific position for ablation
purposes. Under uniform anchoring, the answer is equally likely to
appear at any position in the chain, eliminating the directional bias
of the original sampler.
 
\paragraph{Interaction with the CFG.}
The chain topology is fully determined by the CFG template; the
bidirectional walker does not introduce new relation sequences. It
reads the template as an ordered list of typed slots and fills them
outward from the anchor. For multi-constraint questions, where the
template attaches attribute predicates (e.g., ``whose hobby is
painting'') to interior chain variables, the walker checks these
predicates during the relevant step and resamples if the drawn entity
does not satisfy the constraint.
 
\paragraph{Rejection rate.}
In practice, rejection rates remain below $30\%$ for chains of length
$\le 4$ on graphs with $|V| = 10{,}000$. Beyond length $5$, the
acceptance rate drops and we fall back to a beam over chain prefixes
ranked by candidate-set size.

\subsection{Evaluation Benchmark Details}
\label{app:evaluation_benchmarks}

\begin{enumerate}[itemsep=0.5pt,leftmargin=2em]
    \item \textbf{\hp}~\citep{yang2018hotpotqa}: 2-hop questions from Wikipedia 2018 knowledge graph, requiring composition of information across two passages.
    \item \textbf{\twowiki}~\citep{ho2020constructing}: 2-hop questions built on \hp, spanning four categories: compositional, inference, comparison, and bridge-comparison.
    \item \textbf{\msq}~\citep{trivedi2022musique}: 2--4 hop questions constructed by bridging single-hop questions. We use the Answerable split with gold Wikipedia paragraphs.
    \item \textbf{\synthrm}~\citep{gu2025synthworlds}: A recently introduced, more challenging dataset of 2--6 hop questions derived from January 2025 Wikidata using graph motifs with linear hops and constraints.
    \item \textbf{\synthsm}~\citep{gu2025synthworlds}: A synthetic mirror of \synthrm with identical graph structure but fictional entities. By pairing the two benchmarks, we can isolate compositional reasoning from memorized knowledge.
    \item \textbf{\frames}~\citep{krishna2024frames}: A challenging benchmark with 2--15 hop questions grounded in June 2023 Wikipedia, covering diverse reasoning types including temporal, numerical, and compositional.
\end{enumerate}

\subsection{Comparison Questions of \cofca and \twowiki}
\label{app:comparison_questions}

In \Cref{fig:f1_category_delta_pw_variants_cofca,fig:f1_category_delta_pw_variants_twowiki}, we use the category labels shipped for each question in the \cofca and \twowiki benchmarks: 208 and 119 questions are labeled ``comparison'', respectively (we count the 112 \twowiki bridge-comparison questions as non-comparison).
\cofca contains 223 unlabeled questions as well, so we use the following Python script in \Cref{lst:cofca_categories} that uses regexes to label them as comparison or otherwise.

\begin{lstlisting}[style=pythonstyle, float=tp, caption={Rule-based labeling of unlabeled \cofca questions.}, label={lst:cofca_categories}]
import re

COMPARATIVE_ADJECTIVES = r"(older|younger|earlier|later|taller|bigger|smaller|longer|shorter)\b"

def classify(item: dict) -> str:
    """Return the reasoning category for one unlabeled CofCA question.

    Rules are applied in priority order and taken verbatim from the dataset README, so
    that the counts here stay reproducible against that analysis.

    Args:
        item: a raw minidev.json record, with ``question`` and ``sub_questions``.

    Returns:
        One of ``multi-attribute``, ``intersection``, ``comparison``, ``bridge``.
    """
    question = item["question"].lower()
    sub_questions = [sq.lower() for sq in item.get("sub_questions", [])]
    n = len(sub_questions)

    # 1. Multi-attribute: several independent properties of one entity, no comparison.
    if re.search(r"(what|who|where|when).+and.+(what|who|where|when)", question):
        openings = {" ".join(sq.split()[:2]) for sq in sub_questions}
        if len(openings) == n and not re.search(r"\bor\b", question) and "same" not in question:
            return "multi-attribute"

    # 2. Intersection: one entity satisfying two simultaneous conditions.
    if re.search(r"(who|what).+(and also|who is also|that is also|who was also|also the)", question):
        return "intersection"
    if n >= 2:
        parallel = " ".join(sub_questions[0].split()[:3]) == " ".join(sub_questions[1].split()[:3])
        if parallel and re.search(r"(same person|same individual|same)", " ".join(sub_questions)):
            return "intersection"

    # 3. Comparison: two entities queried on one attribute, then compared.
    if re.search(r"\bor\b", question):
        return "comparison"
    if "between" in question and n <= 3:
        return "comparison"
    if "both" in question:
        return "comparison"
    if re.search(COMPARATIVE_ADJECTIVES, question):
        return "comparison"
    if re.search(r"(who|which).+(first|last)\b", question):
        return "comparison"
    if re.search(r"(same|differ)", question):
        return "comparison"
    for sq in sub_questions:
        if re.search(r"(which|who|is).*(more|less|first|last|same|older|younger|earlier|later|differ)", sq):
            return "comparison"
        if re.search(r"(is|are|were).*(same|both|either|neither)", sq):
            return "comparison"
    if n >= 2 and " ".join(sub_questions[0].split()[:3]) == " ".join(sub_questions[1].split()[:3]):
        return "comparison"

    # 4. Everything else chains through an intermediate entity.
    return "bridge"

def labeled_category(type_field: list[str]) -> str:
    """Return the reasoning category carried by an already-labeled ``type`` list."""
    for tag in type_field:
        if tag in ("comparison", "bridge"):
            return tag
    raise ValueError(f"no reasoning category in {type_field}")
\end{lstlisting}

%% file: images_arxiv/retriever_table_qwen3b.tex
\begin{table}[!h]
\centering
\setlength{\tabcolsep}{3pt}
\begin{tabular}{lcccccc}
\toprule
\shortstack[l]{Evaluation\\Setup} & HotpotQA & 2Wiki & MuSiQue & Synth-RM & Synth-SM & FRAMES \\
\midrule
\multicolumn{7}{l}{Qwen2.5-3B} \\
\quad Paper setup & $40.1 \pm 1.9$ & $30.3 \pm 1.9$ & $21.9 \pm 1.6$ & $17.7 \pm 1.5$ & $9.9 \pm 1.2$ & $12.6 \pm 1.0$ \\
\rowcolor{lightorange} \quad Training setup & $39.8 \pm 2.0$ & $29.3 \pm 1.8$ & $18.2 \pm 1.5$ & $17.6 \pm 1.5$ & $10.7 \pm 1.3$ & $11.9 \pm 0.9$ \\
\midrule
\multicolumn{7}{l}{Trained w/ PhantomEnvs} \\
\quad Paper setup & $61.5 \pm 1.9$ & $56.0 \pm 2.0$ & $41.0 \pm 1.9$ & $35.9 \pm 1.8$ & $34.8 \pm 1.9$ & $27.6 \pm 1.4$ \\
\rowcolor{lightorange} \quad Training setup & $56.9 \pm 1.9$ & $54.3 \pm 2.1$ & $38.8 \pm 2.0$ & $31.5 \pm 1.8$ & $31.5 \pm 1.8$ & $23.4 \pm 1.3$ \\
\midrule
\multicolumn{7}{l}{Trained w/ NQ+HotpotQA} \\
\quad Paper setup & $65.9 \pm 1.9$ & $62.5 \pm 2.1$ & $41.4 \pm 2.0$ & $37.6 \pm 2.0$ & $25.1 \pm 1.9$ & $21.3 \pm 1.3$ \\
\rowcolor{lightorange} \quad Training setup & $67.3 \pm 1.8$ & $60.2 \pm 2.1$ & $37.3 \pm 2.0$ & $33.9 \pm 2.0$ & $25.2 \pm 1.9$ & $20.1 \pm 1.3$ \\
\bottomrule
\addlinespace[2pt]
\end{tabular}
\caption{%
    \textbf{Ablation results on retriever configuration.}
    In the main text, we evaluate every model with a deliberately stronger setup than it trains with (\texttt{Qwen3-Embedding-4B}, 20 turns, 32768 context). Here we re-evaluate the same checkpoints under the training setup (\texttt{e5-base-v2}, 10 turns, 8192 context).
    Other index settings remain fixed.
    The weaker training setup generally lowers every model's score, regardless of how it was trained.
    We report results for one training seed---including the Paper setup rows, which therefore differ slightly from \Cref{tab:f1}---with standard errors over the test sets. The other training seed follows the same trend.
    }
\label{tab:retriever_qwen3b}
\end{table}

%% file: images_arxiv/retriever_table_qwen7b.tex
\begin{table}[!h]
\centering
\setlength{\tabcolsep}{3pt}
\begin{tabular}{lcccccc}
\toprule
\shortstack[l]{Evaluation\\Setup} & HotpotQA & 2Wiki & MuSiQue & Synth-RM & Synth-SM & FRAMES \\
\midrule
\multicolumn{7}{l}{Qwen2.5-7B} \\
\quad Paper setup & $44.2 \pm 2.0$ & $29.1 \pm 1.9$ & $24.5 \pm 1.8$ & $23.6 \pm 1.7$ & $15.6 \pm 1.5$ & $16.2 \pm 1.1$ \\
\rowcolor{lightorange} \quad Training setup & $42.6 \pm 2.0$ & $32.4 \pm 2.0$ & $22.0 \pm 1.7$ & $22.7 \pm 1.7$ & $15.0 \pm 1.5$ & $15.3 \pm 1.1$ \\
\midrule
\multicolumn{7}{l}{Trained w/ PhantomEnvs} \\
\quad Paper setup & $62.2 \pm 1.9$ & $65.0 \pm 2.0$ & $47.1 \pm 2.0$ & $40.3 \pm 1.9$ & $41.8 \pm 1.9$ & $34.9 \pm 1.4$ \\
\rowcolor{lightorange} \quad Training setup & $62.7 \pm 1.9$ & $58.8 \pm 2.1$ & $41.8 \pm 2.0$ & $35.6 \pm 1.9$ & $33.4 \pm 1.9$ & $26.9 \pm 1.4$ \\
\midrule
\multicolumn{7}{l}{Trained w/ NQ+HotpotQA} \\
\quad Paper setup & $74.1 \pm 1.7$ & $71.0 \pm 1.9$ & $51.6 \pm 2.1$ & $43.2 \pm 2.0$ & $33.8 \pm 2.1$ & $30.7 \pm 1.4$ \\
\rowcolor{lightorange} \quad Training setup & $71.3 \pm 1.8$ & $69.4 \pm 2.0$ & $49.7 \pm 2.1$ & $37.7 \pm 2.0$ & $30.4 \pm 2.0$ & $19.4 \pm 1.3$ \\
\bottomrule
\addlinespace[2pt]
\end{tabular}
\caption{%
    \textbf{Ablation results on retriever configuration.}
    In the main text, we evaluate every model with a deliberately stronger setup than it trains with (\texttt{Qwen3-Embedding-4B}, 20 turns, 32768 context). Here we re-evaluate the same checkpoints under the training setup (\texttt{e5-base-v2}, 10 turns, 8192 context).
    Other index settings remain fixed.
    The weaker training setup generally lowers every model's score, regardless of how it was trained.
    We report results for one training seed---including the Paper setup rows, which therefore differ slightly from \Cref{tab:f1}---with standard errors over the test sets. The other training seed follows the same trend.
    }
\label{tab:retriever_qwen7b}
\end{table}

%% file: appendix/02_additional_results.tex
\clearpage
\section{Additional Results}
\label{app:additional_results}

\input{images_arxiv/pooled_table_qwen3b}
\input{images_arxiv/pooled_table_llama3b}

\begin{figure}[!h]
    \centering
    \includegraphics[width=\linewidth]{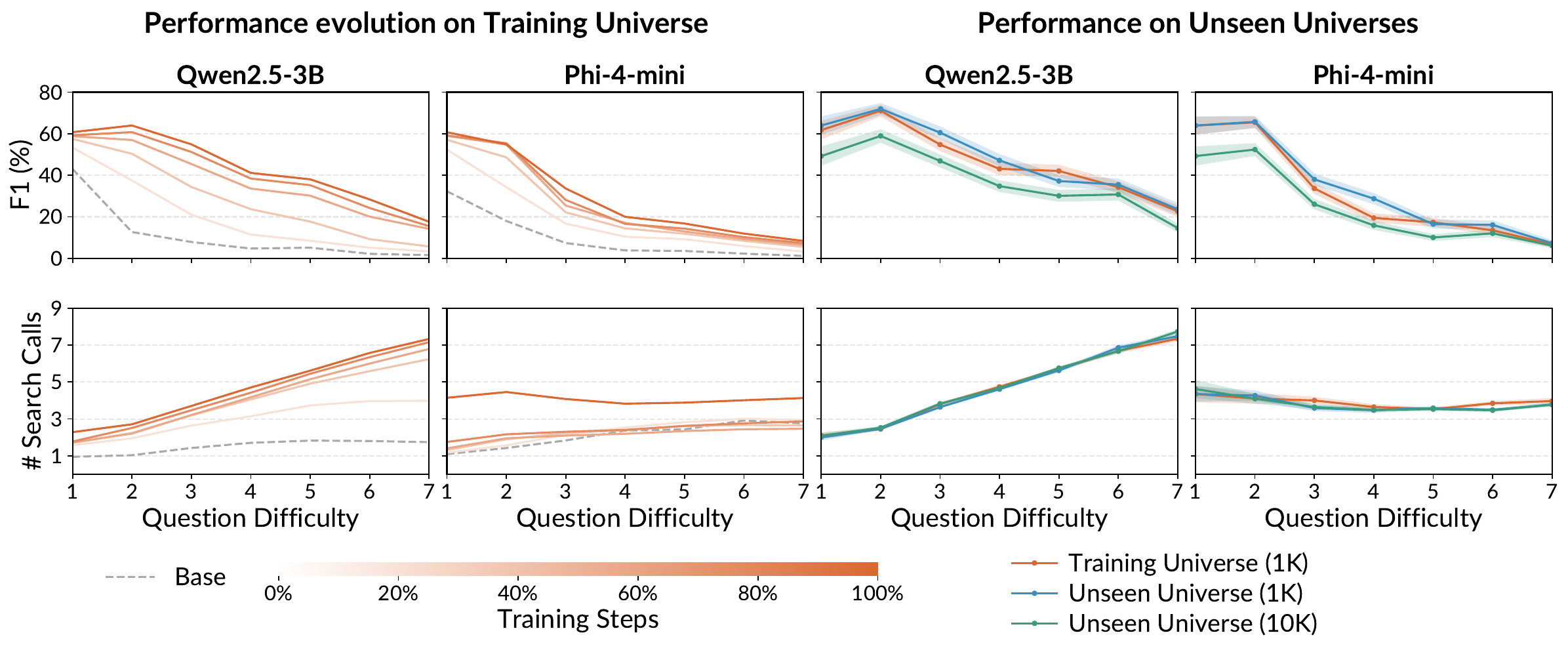}
    \caption{%
    \textbf{F1 scores and number of search calls as a function of question difficulty (hops).}
    We include fine-grained analysis for \qwenthree and \phimini here, see \Cref{fig:pw_generalization_vs_difficulty_qwen7b_llama} and main text for full details.
    }
    \label{fig:pw_generalization_vs_difficulty_qwen3b_phi}
\end{figure}

\begin{figure}[!h]
    \centering
    \includegraphics[width=\linewidth]{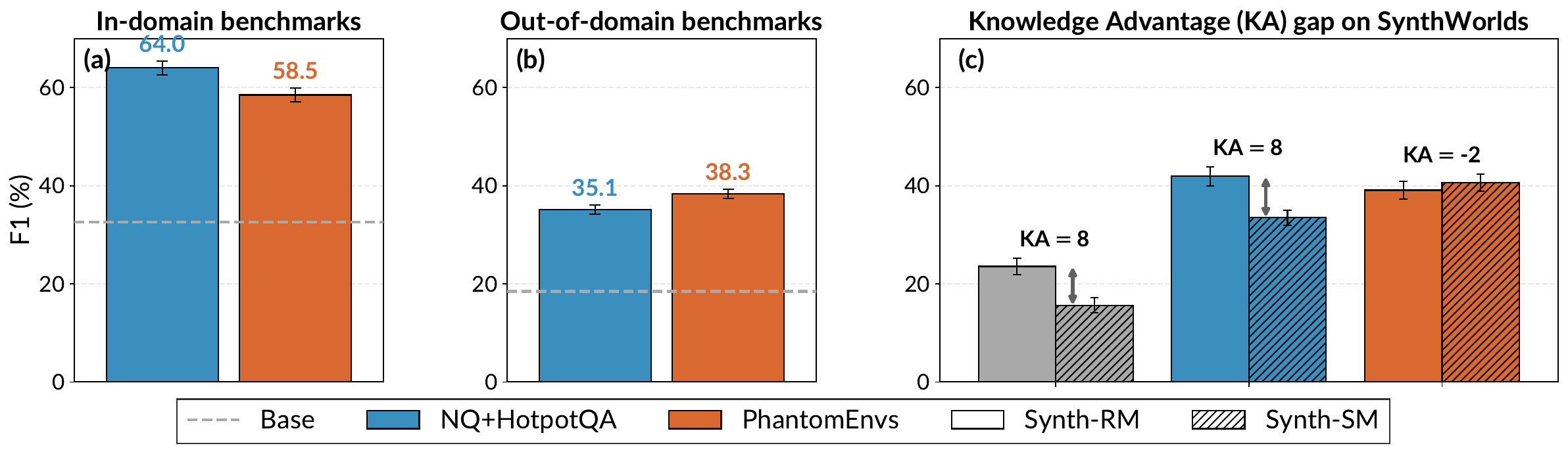}
    \caption{%
    \textbf{\qwenseven performance comparison of real and \penv.}
    See \Cref{fig:f1_transfer_performance_real_vs_pw_qwen3b} and main text for full details.
    }
    \label{fig:f1_transfer_performance_real_vs_pw_qwen7b}
\end{figure}

\begin{figure}[!h]
    \centering
    \includegraphics[width=\linewidth]{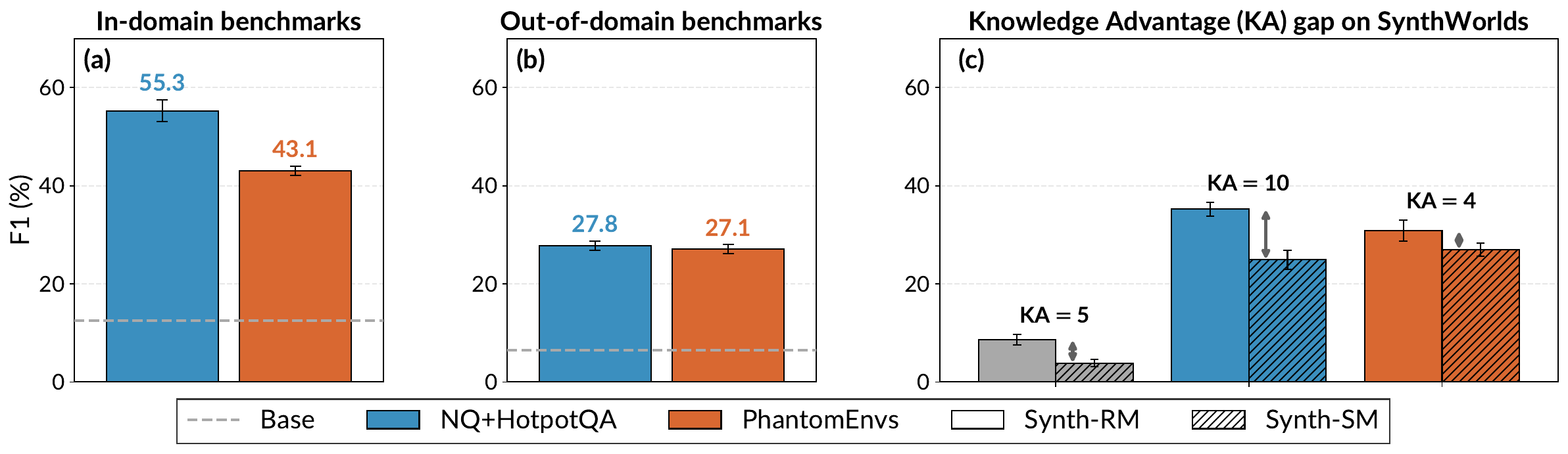}
    \caption{%
    \textbf{\llamathree performance comparison of real and \penv.}
    In out-of-domain benchmarks, \penvshort training reduces but does not fully close the gap to \nqhotpot training for \llamathree, unlike 
    Qwen2.5 models. 
    See \Cref{fig:f1_transfer_performance_real_vs_pw_qwen3b} and main text for full details.
    }
    \label{fig:f1_transfer_performance_real_vs_pw_llama3}
\end{figure}

\begin{figure}[!h]
    \centering
    \includegraphics[width=\linewidth]{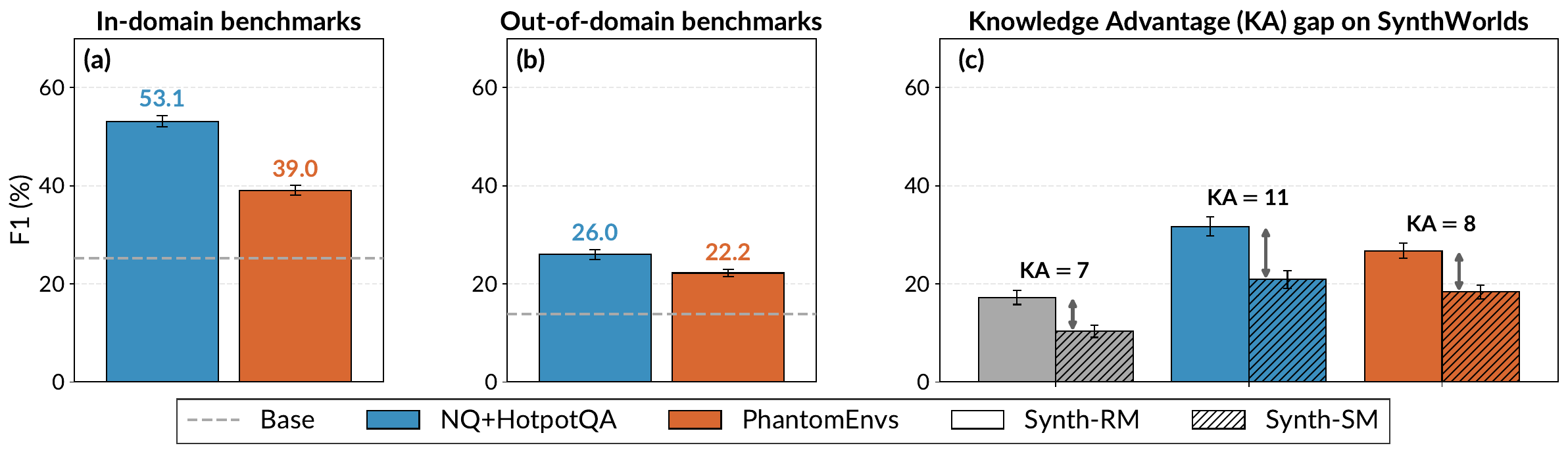}
    \caption{%
    \textbf{\phimini performance comparison of real and \penv.}
    In out-of-domain benchmarks, \penvshort training reduces but does not fully close the gap to \nqhotpot training for \phimini, unlike 
    Qwen2.5 models.
    See \Cref{fig:f1_transfer_performance_real_vs_pw_qwen3b} and main text for full details.
    }
    \label{fig:f1_transfer_performance_real_vs_pw_phi}
\end{figure}

\begin{figure}[!h]
    \centering
    \includegraphics[width=0.8\linewidth]{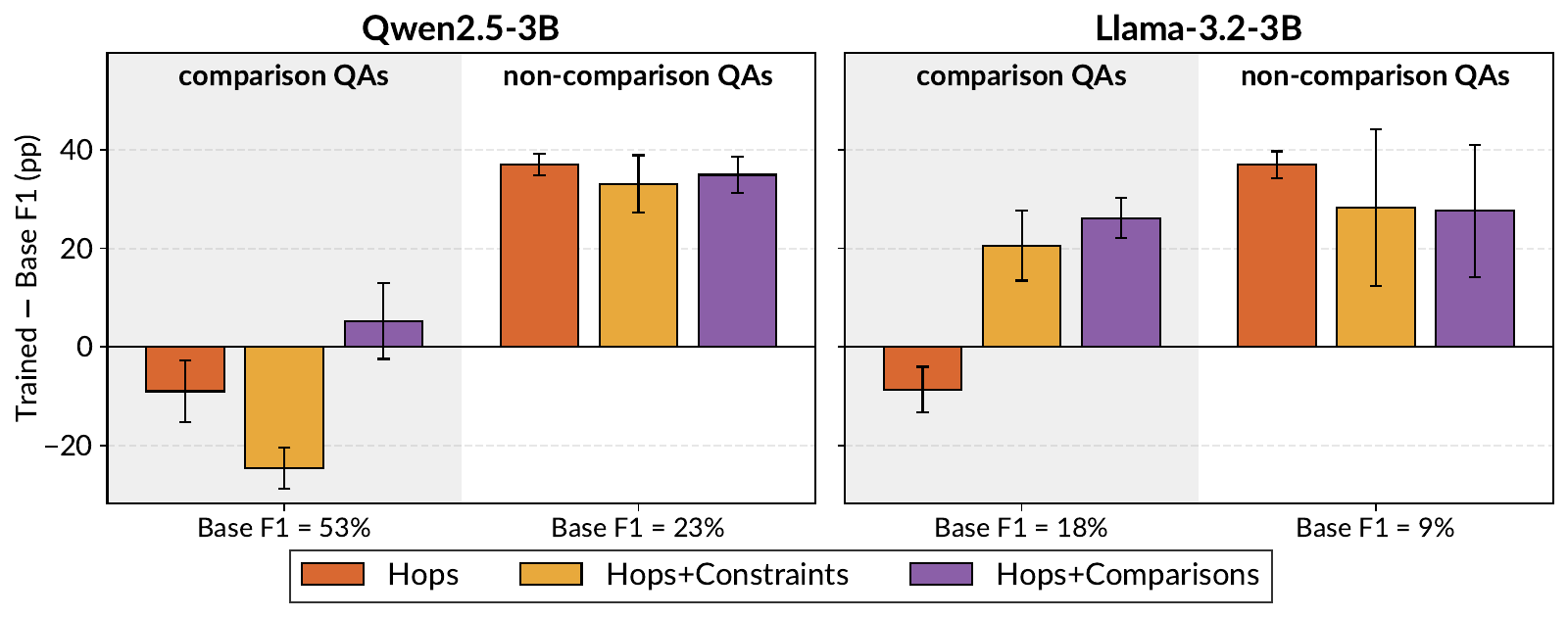}
    \caption{%
    \textbf{F1 score deltas per question type of \twowiki benchmark after fine-tuning on \hops, \hopscomp, and \hopscons environments.}
    We plot F1 score deltas on comparison and non-comparison questions: we calculate F1 score difference per question, average deltas for each category, and report the mean $\pm$ standard error (paired) in percentage points.
    \hopscomp environment improves performance on evaluation comparison questions, but linear hops remains dominant otherwise.
    }
    \label{fig:f1_category_delta_pw_variants_twowiki}
\end{figure}

\input{images_arxiv/nlu_table}

\clearpage
\subsection{Shortcut Trajectories from \hopscons-Trained Agents}
\label{app:hopscons_shortcut_trajectories}

In \Cref{fig:f1_transfer_performance_pw_vs_variants,fig:f1_category_delta_pw_variants_cofca,fig:f1_category_delta_pw_variants_twowiki}, we observed that \hopscons environment showed worse real-world transfer than the other environment complexity axes.
We trace it to a shortcut that the \hopscons environment unintentionally rewards during training.
A constraint such as a specific date of birth or occupation is information-rich: a single dense-retrieval query that pastes the entire question often surfaces the gold document directly, because the rare attribute alone narrows the candidate set.
The agent then never has to resolve the question chain one at a time, and can side-step question decomposition.
Once this works in training, the behavior carries over at test time.
We observe that the \hopscons-trained agent issues the question \textit{verbatim} as its first search on an overwhelming majority of real-world trajectories and almost always paraphrases the same query when retrieval fails.
Whereas, the \hops-trained agent decomposes questions and reasons explicitly between turns, recovering the correct answer more often.
We show one such trajectory pair verbatim below.
Real-world benchmarks rarely include the kind of information-rich attribute filters that make the shortcut work, explaining why \hopscons shows worse real-world transfer.

Here we copy two trajectories on the same question from \msq evaluation---from \qwenthree fine-tuned on \hopscons and on \hops.
The contrast is in how the trained agents answer: the \hopscons agent issues the question as a single monolithic query and only paraphrases it, while the \hops agent decomposes the question and reasons explicitly between turns.
On this instance, only the decomposed trajectory recovers the correct answer.

\textbf{Agent trained on \hopscons (monolithic shortcut, wrong answer).}
The agent issues the question verbatim and then five near-duplicate paraphrases of the same monolithic query.
The retrieved passages name both ``Too Late for Goodbyes'' (Julian Lennon's own song) an ``Hey Jude'' (the McCartney song written for Julian Lennon as a child).
Without decomposing through the bridge entity ``Julian Lennon'', the agent cannot tell which one Paul McCartney wrote, and picks the wrong answer.

\begin{tcolorbox}[colback=gray!5, colframe=gray!40,
                  title={\msq, gold answer: ``Hey Jude''}]
\small
\textbf{Question:} What song did Paul McCartney write for the performer
of Valotte?

\medskip
\texttt{<search>} What song did Paul McCartney write for the performer
of Valotte? \texttt{</search>}

\texttt{<search>} What song did Paul McCartney write for Julian Lennon's
performer Valotte? \texttt{</search>}

\texttt{<search>} What song did Paul McCartney write for Valotte?
\texttt{</search>}

\texttt{<search>} What song did Paul McCartney write for Julian Lennon's
Valotte performance? \texttt{</search>}

\texttt{<search>} What song did Paul McCartney write for Julian Lennon's
Valotte performance? \texttt{</search>}

\texttt{<search>} What song did Paul McCartney write for Julian Lennon's
Valotte performance? \texttt{</search>}

\texttt{<answer>} Too Late for Goodbyes \texttt{</answer>}
\end{tcolorbox}

\textbf{\hops-trained agent (decomposed, correct answer).}
On the same question, the \hops-trained agent first plans the decomposition, then issues sub-questions that resolve the chain left-to-right: it first identifies the performer of Valotte, then asks which song McCartney wrote for that performer.
We show the verbatim reasoning text between retrieval rounds.

\begin{tcolorbox}[colback=gray!5, colframe=gray!40,
                  title={Same question, \hops-trained \qwenthree}]
\small
To find out what song Paul McCartney wrote for the performer of
``Valotte,'' let's break this down into smaller parts.

1. Identify the performer of ``Valotte.''

2. Determine which song Paul McCartney wrote for that performer.

Let's start with finding the performer of ``Valotte.''

\texttt{<search>} Who is the performer of the song ``Valotte''?
\texttt{</search>}

\medskip
\emph{[retrieved doc names Julian Lennon as the artist behind the album
Valotte]}
\medskip

From the search results, it seems there might be a mix-up. The song
``Valotte'' mentioned in the question is likely referring to Julian
Lennon's song ``Too Late for Goodbyes,'' not any of the titles provided
in the search results. Let's verify this by checking if Julian Lennon is
indeed the performer of ``Too Late for Goodbyes.''

\texttt{<search>} Who is the performer of the song ``Too Late for
Goodbyes''? \texttt{</search>}

\medskip
\emph{[retrieved doc confirms Julian Lennon]}
\medskip

From the search results, we confirm that the performer of ``Too Late
for Goodbyes'' is Julian Lennon. Next, we need to find out which song
Paul McCartney wrote for Julian Lennon.

\texttt{<search>} What song did Paul McCartney write for Julian Lennon?
\texttt{</search>}

\medskip
\emph{[retrieved doc explains ``Hey Jude'' was written by McCartney to
comfort Julian Lennon during his parents' divorce]}
\medskip

From the search results, we find that one of the songs Paul McCartney
wrote for Julian Lennon is ``Hey Jude.''

Therefore, the song Paul McCartney wrote for the performer of
``Valotte'' (which we determined to be Julian Lennon) is
\texttt{<answer>} Hey Jude \texttt{</answer>}.
\end{tcolorbox}

%% file: images_arxiv/pooled_table_qwen3b.tex
\begin{table}[!htbp]
\centering
\setlength{\tabcolsep}{6pt}
\begin{tabular}{lcc}
\toprule
Corpus & Qwen2.5-3B & + PhantomEnvs \\
\midrule
Benchmark's own & $22.1 \pm 0.6$ & $42.1 \pm 0.6$ \\
\rowcolor{lightorange} All pooled & $21.2 \pm 0.6$ & $40.7 \pm 0.7$ \\
\bottomrule
\addlinespace[2pt]
\end{tabular}
\caption{%
    \textbf{Average F1 scores of \qwenthree, ablating choice of search corpus.}
    }
\label{tab:pooled_qwen3b}
\end{table}

%% file: images_arxiv/pooled_table_llama3b.tex
\begin{table}[!htbp]
\centering
\setlength{\tabcolsep}{6pt}
\begin{tabular}{lcc}
\toprule
Corpus & Llama-3.2-3B & + PhantomEnvs \\
\midrule
Benchmark's own & $9.5 \pm 0.4$ & $35.1 \pm 0.7$ \\
\rowcolor{lightorange} All pooled & $8.1 \pm 0.4$ & $29.9 \pm 0.6$ \\
\bottomrule
\addlinespace[2pt]
\end{tabular}
\caption{%
    \textbf{Average F1 scores of \llamathree, ablating choice of search corpus.}
    }
\label{tab:pooled_llama3b}
\end{table}

%% file: images_arxiv/nlu_table.tex
\begin{table}[!htbp]
\centering
\setlength{\tabcolsep}{6pt}
\begin{tabular}{lccc}
\toprule
Model & PopQA SubEM (\%, $\uparrow$) & MMLU Acc (\%, $\uparrow$) & Wiki-18 PPL ($\downarrow$) \\
\midrule
Qwen2.5-3B & $14.15 \pm 0.29$ & $66.37 \pm 0.38$ & $13.82 \pm 0.01$ \\
\rowcolor{lightorange} \quad + PhantomEnvs & $14.39 \pm 0.21$ & $66.51 \pm 0.28$ & $13.94 \pm 0.01$ \\
\rowcolor{lightorange} \quad + NQ+HotpotQA & $14.78 \pm 0.21$ & $66.38 \pm 0.27$ & $13.88 \pm 0.01$ \\
\midrule
Qwen2.5-7B & $16.51 \pm 0.31$ & $74.27 \pm 0.35$ & $12.60 \pm 0.01$ \\
\rowcolor{lightorange} \quad + PhantomEnvs & $16.65 \pm 0.23$ & $74.31 \pm 0.25$ & $12.79 \pm 0.02$ \\
\rowcolor{lightorange} \quad + NQ+HotpotQA & $16.44 \pm 0.24$ & $74.18 \pm 0.25$ & $12.65 \pm 0.04$ \\
\midrule
Llama-3.2-3B & $14.75 \pm 0.30$ & $60.58 \pm 0.40$ & $17.42 \pm 0.01$ \\
\rowcolor{lightorange} \quad + PhantomEnvs & $14.87 \pm 0.22$ & $60.45 \pm 0.28$ & $17.66 \pm 0.02$ \\
\rowcolor{lightorange} \quad + NQ+HotpotQA & $13.98 \pm 0.21$ & $60.28 \pm 0.28$ & $17.73 \pm 0.01$ \\
\midrule
Phi-4-mini & $17.59 \pm 0.32$ & $68.69 \pm 0.37$ & $16.74 \pm 0.01$ \\
\rowcolor{lightorange} \quad + PhantomEnvs & $15.50 \pm 0.30$ & $68.49 \pm 0.37$ & $17.87 \pm 0.01$ \\
\rowcolor{lightorange} \quad + NQ+HotpotQA & $16.86 \pm 0.31$ & $68.56 \pm 0.37$ & $16.89 \pm 0.01$ \\
\bottomrule
\addlinespace[2pt]
\end{tabular}
\caption{%
    \textbf{RL fine-tuning leaves pretrained knowledge and fluency intact.}
    We measure parametric factual recall (PopQA SubEM metric~\citep{mallen2023popqa}), general knowledge and reasoning (MMLU accuracy~\citep{hendrycks2020measuring}), and language-modeling fluency on Wikipedia-2018 (perplexity, 1M passages randomly sampled from the 21M released by~\citep{jin2025searchr1}).
    PopQA does not degrade after fine-tuning (either synthetic or NQ+HotpotQA training), except Phi-4-mini-instruct + \penvshort ($-2.1$ pp).
    MMLU accuracy is unchanged throughout after fine-tuning.
    Wiki-18 perplexity rises slightly after fine-tuning (1-2\% relative, and 7\% for Phi-4-mini-instruct + \penvshort).
    Learning agentic search through RL therefore largely teaches procedural skill, one that can be learned in complement to the LLMs' memorized parametric knowledge.
    This is in line with~\citet{chen2025retaining}, who show that RL fine-tuning can retain parametric knowledge.
    }
\label{tab:nlu}
\end{table}